# Automatic Text Summarization Methods: A Comprehensive Review

Divakar Yadav[1], Jalpa Desai, Arun Kumar Yadav[2]

20mcs105@nith.ac.in, dsy99@rediffmail.com, ayadav@nith.ac.in

ORCID: [1]https://orcid.org/0000-0001-6051-479X , [2]https://orcid.org/0000-0001-9774-7917

**Abstract:**

One of the most pressing issues that have arisen due to the rapid growth of the Internet is known as information overloading. Simplifying the relevant information in the form of a summary will assist many people because the material on any topic is plentiful on the Internet. Manually summarising massive amounts of text is quite challenging for humans. So, it has increased the need for more complex and powerful summarizers. Researchers have been trying to improve approaches for creating summaries since the 1950s, such that the machine-generated summary matches the human-created summary. This study provides a detailed state-of-the-art analysis of text summarization concepts such as summarization approaches, techniques used, standard datasets, evaluation metrics and future scopes for research. The most commonly accepted approaches are extractive and abstractive, studied in detail in this work. Evaluating the summary and increasing the development of reusable resources and infrastructure aids in comparing and replicating findings, adding competition to improve the outcomes. Different evaluation methods of generated summaries are also discussed in this study. Finally, at the end of this study, several challenges and research opportunities related to text summarization research are mentioned that may be useful for potential researchers working in this area.

## 1. Introduction:

The task of compressing a piece of text into a shorter version, minimizing the size of the original text while keeping crucial informational aspects and content meaning, is known as summarization. Fig. 1 shows task of summarization in a simple way. A summary is a reductive transformation of a source text into a summary text by extraction or generation (Radev et al., 2004). According to another definition, "An automatic summary is a text generated by a software that is coherent and contains a significant amount of relevant information from the source text. Its compression rate τ is less than a third of the length of the original document (Hovy & Lin, 1996). The ratio between the length of the summary and the length of the source document is calculated by the compression rate τ as shown below:

$$\tau = \frac{|Summary|}{|Source|}$$

Where | • | indicates the length of the document in characters, words, or Sentences. τ can be expressed as a percentage. In fact, (C. Y. Lin, 1999) study shows that the best performances of automatic summarization systems are found with a compression rate of τ = 15 to 30% of the length of the source document.

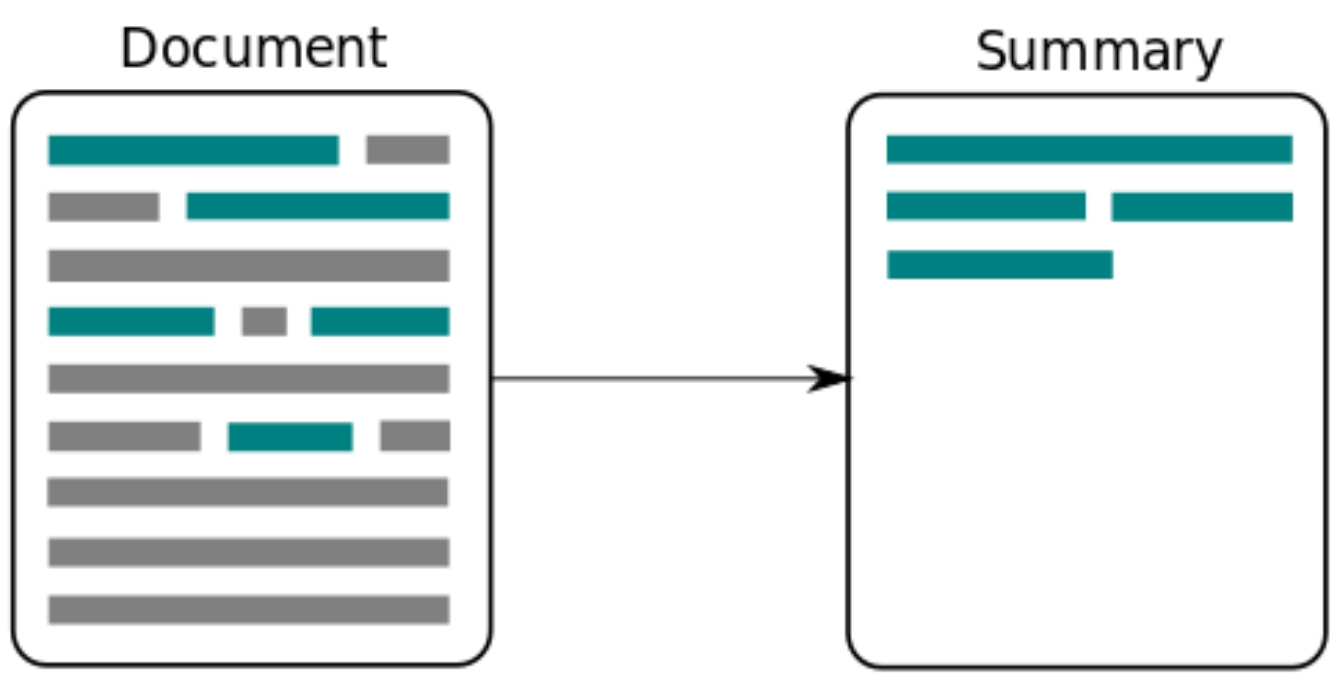

Fig. 1: generating summary from input document

Understanding the source text and creating a brief and abbreviated version of it are two processes in the human generation of summaries. Fig. 2 shows how human produces the summaries of an original text document. The summarizer's linguistic and extra-linguistic abilities and knowledge are required for both understanding the material and producing summaries. Although people can write better summaries (in terms of readability, content, form, and conciseness). Automatic text summarizing is a useful supplement to manual summation rather than a substitute.

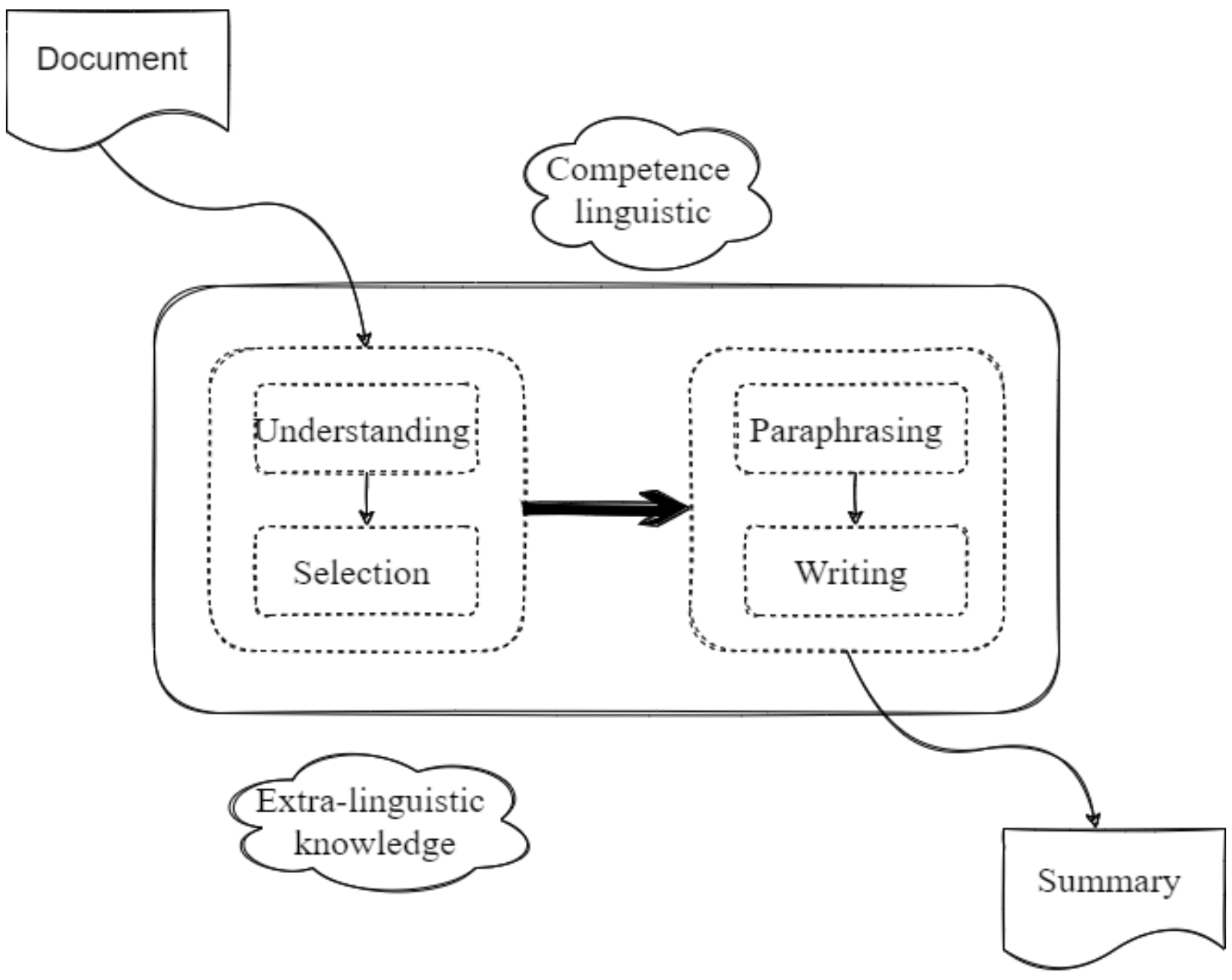

Fig. 2: human's process for generating summaries

### 1.1 Requirement of Text Summarization

The adage "too much information kills information" is as relevant today as it has ever been. The fact that the Internet is available in various languages only adds to the aforementioned document analysis challenges. Automatic text summarization aids in the effective processing of an ever-increasing volume of data that humans are just unable to handle. Let's look at some eye-opening facts about the world of data provided by Arne von See (2021) as shown in fig. 3. Some facts about it are: in the previous two years, 90 percent of the world's data has been created. The majority of businesses only look at 12% of their data. Each year, bad data costs the United States $3.1 trillion. By 2025, the amount of data created will have surpassed 180 zettabytes. To download all of the material from the internet now, it would take a human around 181 million years.

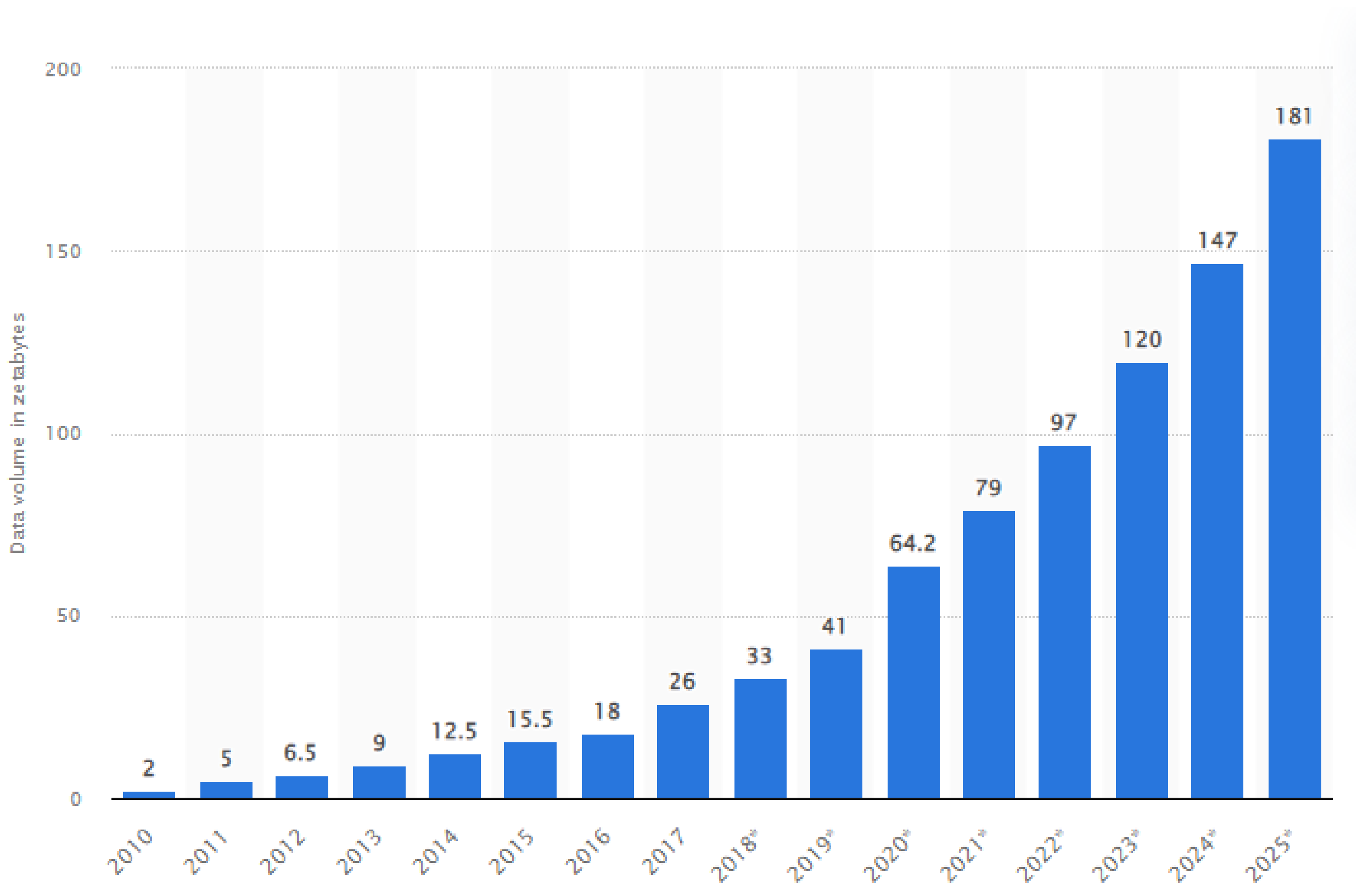

Fig. 3. Volume of data/information created, captured, copied, and consumed worldwide from 2010 to 2025 (Arne von See, 2021)

There are several valid reasons in favour of the automatic summarization of documents. Here are listed just a few (Ab & Sunitha, 2013)

i. Summaries saves reading time.
ii. Summaries help in the selection of documents when conducting research.
iii. Indexing is more successful when automatic summarization is used.
iv. When compared to human summarizers, automatic summary systems are less biased.
v. Because they provide personalized information, personalized summaries are important in question-answering systems.
vi. Commercial abstract services can improve the number of texts they can process by using automatic or semi-automatic summarizing techniques.

Automatic Text Summarization (ATS) is a relatively new learning issue that has gotten a lot of interest. As research advances, we hope to see a breakthrough that will help with this by giving a timely technique of summarising big texts. We present an overview of text summarising techniques in this work to highlight their usefulness in dealing with enormous data and to assist researchers in using them to address challenges**.** The fig. 4 shows the number of researcher papers published in domain of text summarization in a particular time interval staring from 1958.

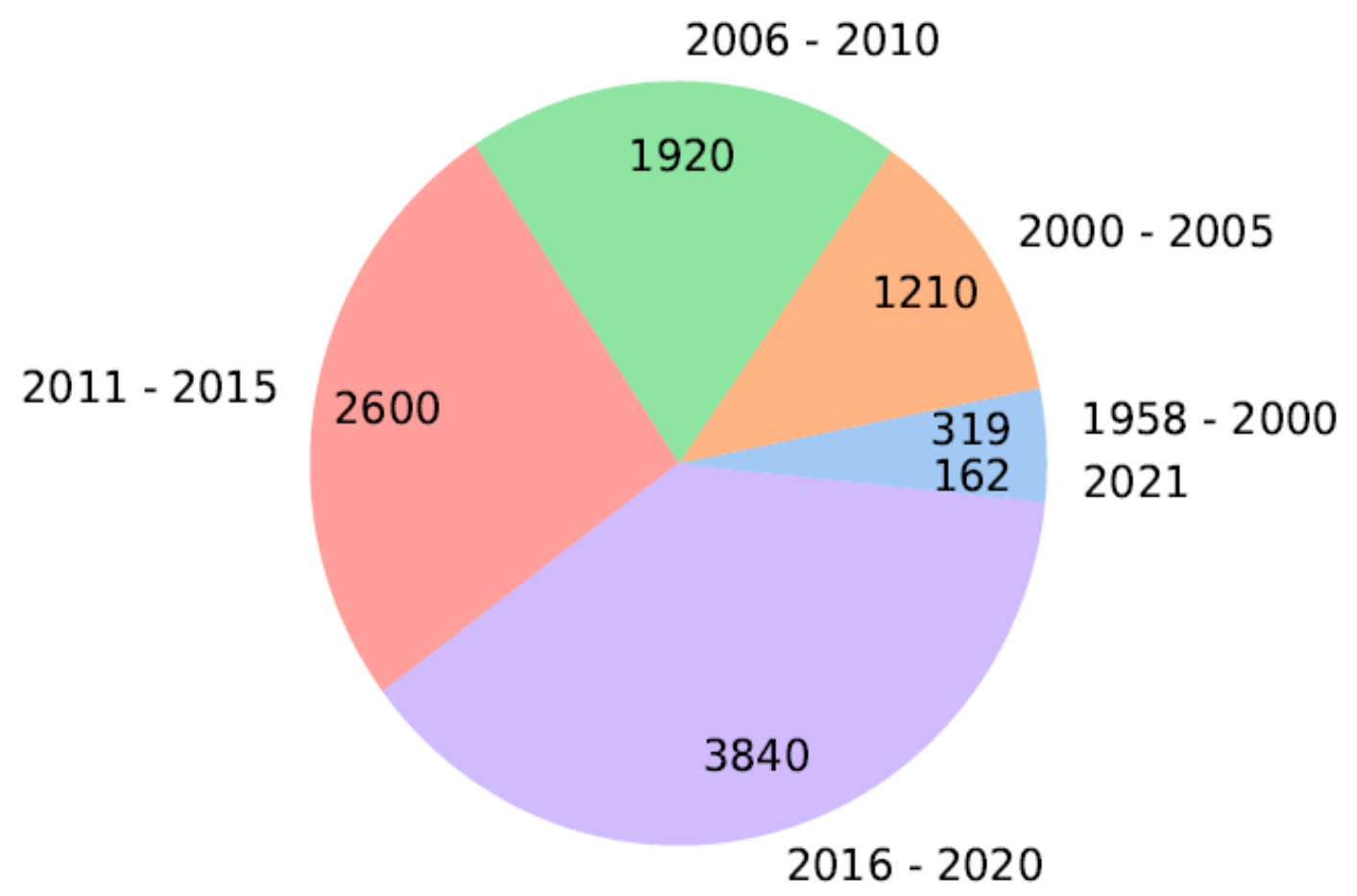

Fig. 4: Number of Research articles published in the domain of ATS in different time interval

**1.2 Main Contribution of study**

This work provides a concise, current and comprehensible view in the field of text summarization. The major contributions of this study are as under:

a. Starting from ground level, make the reader comfortable with the ATS system and why we need an ATS system. Provided examples of ATS systems presented in the literature for each application and illustrated ATS systems' classifications.
b. Provided a detailed analysis of the three ATS approaches extractive, abstractive, and hybrid. Furthermore, the review table is built on factors like dataset, approach, performance, advantages, and disadvantages.
c. Provided an overview of the standard datasets and provided complete details about evaluation methods available for the ATS system.
d. Detailed analysis of challenges and future scopes for text summarization.

This article is arranged into six sections. Section 1 discusses the introduction of an automatic text summarization system with its requirements and applications. The automatic text summarization is divided into many categories discussed in detail in section 2. Next, section 3 focused on Extractive, Abstractive and Hybrid text summarization. The evaluation methods for summaries generated by the system are discussed in section 4. Later that frequently used datasets for summarization task is listed in section 5. Lastly, the conclusion is given in section 6.

## 2. Categorization of ATS

There are different classifications for an automatic text summarization (ATS) system based on its input, output, purpose, length, algorithms, domain, and language. There are many other factors that can be considered while discussing the classification of summarization. Different researchers have considered different factors. As per our survey, the detailed categorization of an ATS system is given in fig. 5. A detailed explanation of a particular category are discussed in following sub-sections as under:

### 2.1 Based on no. of Input documents

Based upon size of input source documents that are used to generate a summary, summarization can be divided in two types:

- **Single Document**: Single document text summarization is automatic summarization of information a single document (Garner, 1982).

- **Multiple Document:** Multi-document text summarization is an automatic summarization of information from multiple document (Ferreira et al., 2014).

Multi-document summarization is important where we must put different types of opinions together, and each idea is written with multiple perspectives within a single document. Single document text summarization is easy to implement, but multi-document summarization is a complex task. Redundancy is one of the biggest problems in summarizing multiple documents. Carbonell & Goldstein (1998) has given MMR (Maximal Marginal Relevance) approach, which helps to reduce redundancy. Another main problem for multi-document summarization is heterogeneity within a large set of documents. It is very complex to summarize multiple documents with extractive methods where there are so many conflicts and biases in the real world. Here for multiple documents, abstractive summarization performs far better. However, multi-document summarization also brings issues like redundancy in output summary while working with a huge number of documents. Single document text summarization is used in a limited field like reading the given comprehension and giving an appropriate title or summary. In contrast, multi-document text summarization can be used in the field of news summarization from different sites, customer's product reviews from different vendors, Q&A systems and many more.

SummCoder (Joshi et al., 2019) is a new methodology for generic extractive single document text summarization. The method creates a summary based on three criteria they developed: sentence content relevance, sentence novelty, and sentence position relevance. The novelty metric is produced by utilizing the similarity among sentences represented as embedding in a distributed semantic space, and the sentence content relevance is assessed using a deep auto-encoder network. The sentence position relevance metric is a custom feature that gives the initial few phrases more weight thanks to a dynamic weight calculation method controlled by the document length. In the extractive multi-document text summarization field, Sanchez-Gomez et al. (2021) shows that all feasible combinations of the most prevalent term-weighting schemes and similarity metrics have been implemented, compared, and assessed. Experiments with DUC datasets were conducted, and the model's performance was evaluated using eight ROUGE indicators and the execution time. The TF-IDF weighting scheme and cosine similarity give the best result of 87.5% ROUGE score as a combination.

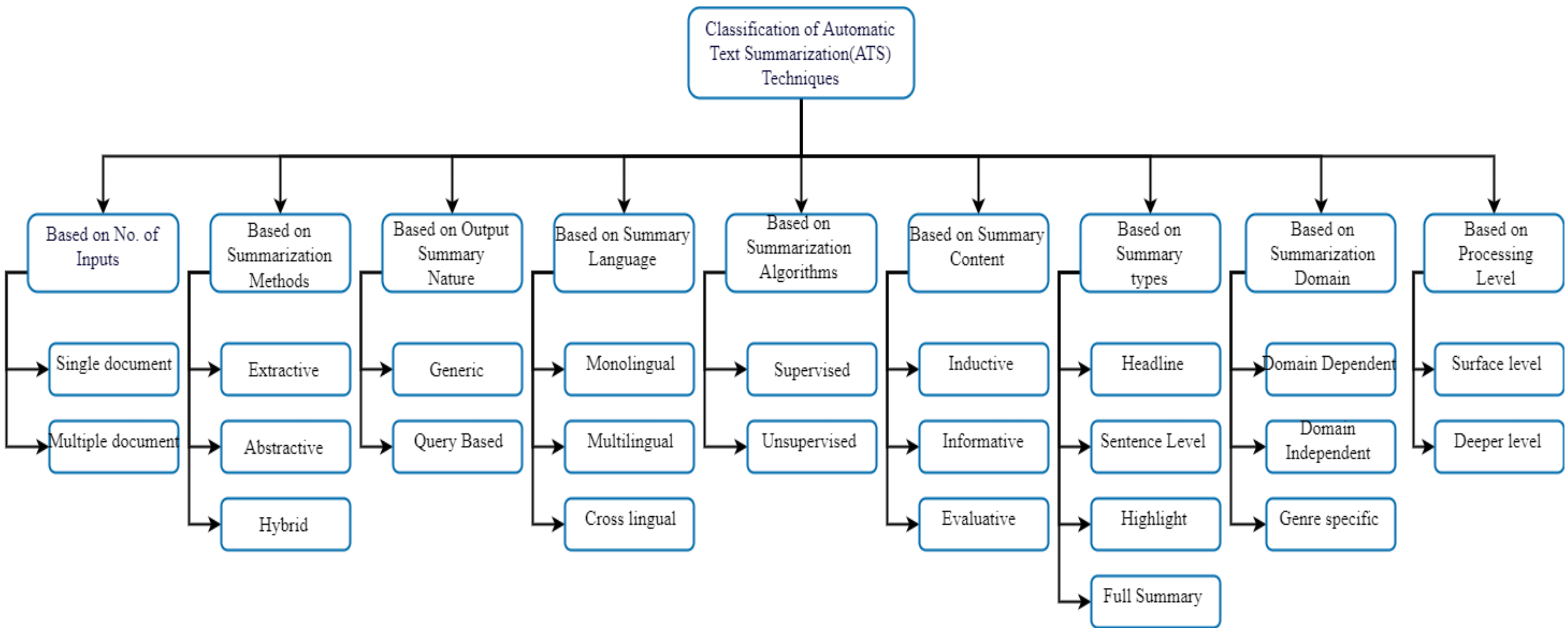

Fig. 5: Detailed Categorization of automatic text summarization system

## 2.2 Based on Summarization Methods

Based on methods that how can summaries are produced, i.e. Just picking up sentences from the source text or generating new sentences after reading source text or a combination of both, summarization can be divided into three types:

- ***Extractive Automatic Text Summarization:*** Extractive text summarization is the strategy of concatenating on extracting summary from a given corpus (Rau et al., 1989).
- ***Abstractive Automatic Text Summarization:*** Abstractive text summarization involves paraphrasing the given corpus and generating new sentences (Zhang et al., 2019).
- ***Hybrid Automatic Text Summarization:*** It combines both extractive and abstractive methods. It means extracting some sentences and generating a new one from a given corpus (Binwahlan et al., 2010).

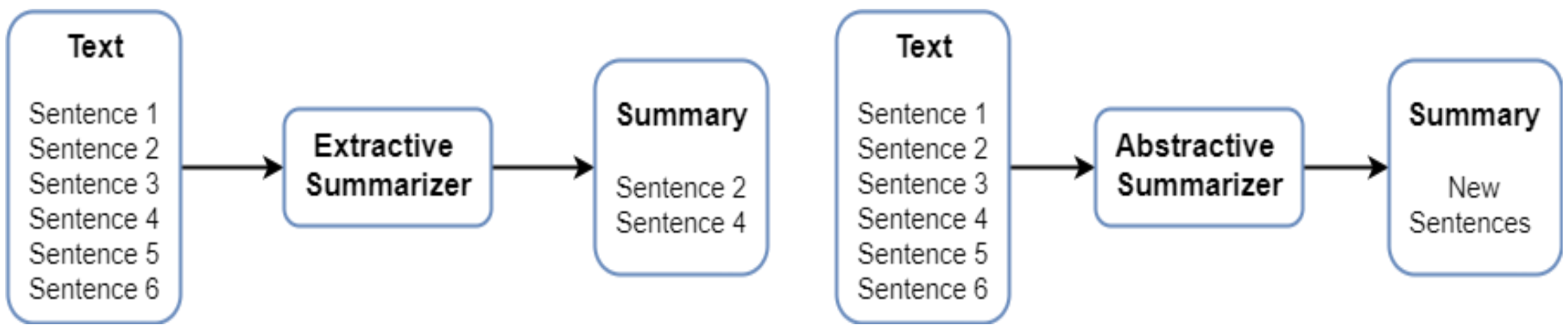

Fig. 6: Extractive text summarizer and Abstractive text summarizer

Think of a highlighter used to point out important sentences in a book. It is an example of extractive text summarization. Now think of the notes we prepare from a book using our own words. It is an example of abstractive text summarization. Extractive text summarization is like copy-pasting some of the important sentences from the source text, while abstractive text summarization selects some meaningful sentences and generates new sentences from previously selected sentences. Refer Fig. 6 for a better understanding of Extractive and Abstractive summarization. Hybrid text summarization combines an approach for producing a summary efficiently. Both Extractive and Abstractive text summarization falls into Machine Learning and NLP domain. Additionally, abstractive text summarization covers NLG. The survey of both approaches is shown in the later section of this article. Critical areas where extractive text summarization is applied are news, medical, book, legal document, abstractive text summarization, customer reviews, blog, tweet summarization, etc.

The plus point of the extractive text summarization model is that the sentences in the summaries must adhere to the syntactic structure's constraints. However, that model's shortcoming is that the summaries' sentences may not be semantically meaningful. This disadvantage arises because adjacent sentences in the summaries are not always contiguous in the original text. Because ATS models learn the collocation between words and construct a sequence of keywords based on the collocation between words after training, they have the advantage of inclusive semantics. The downside of ATS models is that it is challenging to meet the criterion of syntactic structure with this sequence of keywords. Rare words are another major flaw in traditional ATS models. The number of occurrences of a rare word and its collocation will define its importance, but humans will use other elements to assess whether a word is essential. As a result, in some instances, some words that appear infrequently might be deemed unimportant, although a portion of these words is critical for summary construction from a human perspective (Song et al., 2019b).

## 2.3 Based on Output Summary Nature:

Based on the output summary's characteristics the ATS system can be divided into two types:

- ***Generic:*** Generic text summarizers fetch important information from one or more documents to provide a concise meaning of given document(s) (Aone et al., 1997).

- ***Query-Based:*** A query-based summarizer is built to handle multi-documents and gives a solution as per the user's query (Van Lierde & Chow, 2019). The score of sentences in each document is based on the frequency counts of words or phrases in query-based text summarization. Sentences containing query phrases receive higher marks than sentences containing single query words. The sentences with the highest scores and their structural context are extracted for the output summary (Kiyani & Tas, 2017)..

A query-based sentence extraction algorithm is given as below (Pembe & Güngör, 2007):

i. Rank all the sentences according to their score.
ii. Add the main title of the document to the summary.
iii. Add the first level-1 heading to the summary.
iv. While (summary size limit not exceeded)
v. Add the next highest scored sentence.
vi. Add the structural context of the sentence: (if any and not already included in the summary)
vii. Add the highest-level heading above the extracted text (call this heading h).
viii. Add the heading before h in the same level.
ix. Add the heading after h in the same level.
x. Repeat steps 7, 8 and 9 for the subsequent highest-level headings.
xi. End while

A query is not used in generic summaries. Because they do not comprehensively assess the original document, query-based summaries are biased. They are not ideal for content overview because they solely deal with user queries. Generic summaries are necessary to specify the document's category and to describe the document's essential points. The key subjects of the documents are considered in the best general summary, which strives to minimize redundancy as much as possible (Kiyani & Tas, 2017).

### 2.4 Based on Summary Language

Based on the language of input and output of the ATS system, it can be divided into the following 3 categories:

- ***Monolingual:*** In a Monolingual text summarizer, the language of the input document and output summary is the same (Kutlu et al., 2010).
- ***Multilingual:*** In a Multilingual text summarizer, input is written in many languages (Hindi, English, and Gujarati), and output summary is generated likewise in these languages (Hovy & Lin, 1996).
- ***Cross- Lingual:*** In a Cross-lingual text summarizer, the input document is in one language (say English), and the output summary is in another language (say Hindi) (Linhares Pontes et al., 2020).

Most of the research papers studied in this article are based on monolingual text summarization. Compared to monolingual, multilingual and cross-lingual is challenging to implement. It takes more effort to train a machine on more than one language structure. SUMMARIST (Hovy & Lin, 1996) is a multilingual text summarization system based on an extraction strategy that generates summaries from English, Indonesian, Spanish, German, Japanese, Korean, and French sources. Cross-Language Text Summarization (CLTS) (Linhares Pontes et al., 2020) generates a summary in a target language from source language materials. It entails a combination of text summarising and machine translation methods. Unfortunately, this combination leads to mistakes, which lowers the quality of summaries. CLTS systems may extract relevant information from both source and destination languages through joint analysis, which improves the development of extractive cross-lingual summaries. Recent methods for CLTS have offered compressive and abstractive approaches; however, these methods rely on frameworks or tools that are only available in a few languages, restricting their applicability to other languages.

### 2.4 Based on Summarization Algorithms

Based on the actual algorithm that is used to generate the summarises, the ATS system is divided into two types as given below:

- ***Supervised:*** The supervised summarizer needs to train the sample data by labelling the input text document with the help of human efforts (Neto et al., 2002).
- ***Unsupervised:*** In the Unsupervised summarizer training phase is not needed (Alami et al., 2019).

In order to select important content from documents in a supervised system, training data is required. Training data is a large volume of labelled or annotated data is required for learning techniques. These systems are approached as a two-class classification issue at the sentence level, with positive samples being sentences that belong to the summary and negative samples being sentences that do not belong to the summary. On the other hand, unsupervised systems do not require any training data. They create the summary by just looking at the documents they want to look at for summarization. As a result, they can be used with any newly observed data without needing extra adjustments. These systems use heuristic methods to extract relevant sentences and construct a summary. Clustering is used in unsupervised systems (Gambhir & Gupta, 2017).

A single document supervised machine learning-based approach for the Hindi language is given by Nikita (2016). The sentences are divided into four categories: most influential, important, less important, and insignificant. The summarizer is then trained using the SVM supervised machine learning algorithm to extract important sentences based on the feature vector. Sentences are then included in the final summary based on the required compression ratio. The experiment was carried out on news stories from various categories such as Bollywood, politics, and sports, and the results showed 72 percent accuracy at a compression ratio of 50 percent and 60 percent at a compression ratio of 25 percent. Recently, an unsupervised neural network approach has been studied by Meknassi et al. (2021) for Arabic language text summarization. A new approach using documents clustering, topic modelling, and unsupervised neural networks have been proposed to build an efficient document representation model to overcome problems raised with Arabic text documents. The proposed approach is evaluated on Essex Arabic Summaries Corpus and compared against other Arabic text summarization approaches using ROUGE measure.

### 2.5 Based on Summary Content

Based on the type of the content of the output summaries, the system is categorised into two parts as:

- ***Inductive:*** Indicative summary contains only generic idea about the source document (Bhat et al., 2018).
- ***Informative:*** Informative summary contains all the main topics about the original document (Bhat et al., 2018)**.**
- ***Evaluative or Critical:*** It capture the summary from author's point of view on a given topic (Jezek & Steinberger, 2008)**.**

Inductive summarization is used to indicate what the document is all about, and it aims to give an idea to a user whether to read this original document or not. The length of this summary is approximately 5% of the original content. The informative summarization system summarises the primary text concisely. The helpful summary is around 20% of the whole text length (Kiyani & Tas, 2017). A typical example of evaluative summaries are reviews, but they are pretty out of the scope of nowadays summarizers. It should be emphasized that the three groupings indicated above are not mutually exclusive and are common summaries that have both an informative and an indicative role. Informative summarizers are frequently used as a subset of indicative summarizers (Jezek & Steinberger, 2008).

### 2.6 Based on Summary types

Based on the length of the generated summaries, the ATS system is divided into four types as given below:

- ***Headline:*** A headline generated from a source document is usually shorter than a sentence (Barzilay & Mckeown, 2005).
- ***Sentence level:*** The sentence-level summarizer produces a single sentence from the original text (Y. H. Hu et al., 2017).
- ***Highlight:*** Highlights are produced in a compressed form of the original text written in bullet points (Tomek, 1998).
- ***Full summary:*** Full summary is generated as per the user's compression rate or user's requirements (Koupaee & Wang, 2018)

Headlines, highlights, and sentence level type of summary are generally used in news database or opinion mining or for social media dataset whereas a full summary is commonly used for all the domains.

### 2.7 Based on Summarization Domain:

Based on the domain of the input and output of the ATS system, it is divided into following 3 categories:

- ***Genre Specific:*** It accepts only special type of input text format (Hovy & Lin, 1996) .
- ***Domain dependent:*** Domain dependent summarization is specific to one domain (Farzindar & Lapalme, 2004)**.**
- ***Domain independent:*** Domain independent summarization system is independent of source documents' domain.

In genre-specific summarization, there is a restriction on the text template. Newspaper articles, scientific papers, stories, instructions, and other types of templates are available. The summary is generated by the system using the structure of these templates. On the other hand, independent systems have no predefined limitations and can take a variety of text kinds. Furthermore, some techniques only summarise texts whose subject can be characterized in the system's domain; these systems are domain-dependent. These systems impose some restrictions on the topic matter of documents. Such systems know everything there is to know about a specific subject and use that knowledge to summarise. Generally, graph-based techniques are adopted for domain-dependent summarisation as they have sound potential. The authors of (Moradi et al., 2020) have given an efficient solution to deal with the challenges in graph-based methods. To capture the linguistic, semantic, and contextual relationships between the sentences, they trained the model by continuous word representation model. i.e., Word2vec's Skiagrams and Continuous Bag of Words (CBOW) models(Mikolov et al., 2013) and Global Vectors for Word Representation (GloVe) (Mutlu et al., 2020)(Hanson Er, 1971) on a large corpus of biomedical text. To solve the challenge of ranking the most important nodes in a graph, they adopted undirected and weighted graph ranking techniques like the PageRank algorithm (Brin & Page, 2012). Newspaper stories and scientific text have distinct qualities than legal text. In a comprehensive document of the news genre, for example, there is little or no structure. The presence of the same term at different levels of the hierarchy will have distinct effects. The relevance of the words in a judgement is determined by the source of the ruling (whether it is from a District Court, State Court, Supreme Court, or Federal Court). We can generally ignore references/citations when summarizing content; however, this may not be possible in legal writings (Kanapala et al., 2019).

### 2.8 Based on Processing Level

Based on the processing level of the input document, the system is divided into two types:

- ***Surface-level approaches:*** In this scenario, data is represented by shallow feature ideas and their combinations.

Statistically salient terms, positionally salient terms, cue phrases, domain-specific or a user's query terms are examples of shallow features. The outcomes are in the form of extracts (Ježek et al., 2007).

- ***Deeper-level approaches:*** Extracts or abstracts may be produced through deeper-level techniques. In the latter scenario, synthesis is used to generate natural language. It requires some semantic analysis. For example, entity techniques can construct a representation of text entities (text units) and their relationships to identify salient areas. Entity relationships include thesaural, syntactic, and semantic relationships, among others. They can also use discourse methodologies to represent the text structure, such as hypertext mark-up or rhetorical structure (Ježek et al., 2007).

## 3 Detailed about ETS, ABS and HTS

Among all the Automatic text summarization system classification, the most commonly accepted or used categories are Extractive, Abstractive and hybrid summarization. Thus, this article focused is on these two approaches mainly. Now lets us see a detailed survey on these approaches:

### 3.1 Extractive text summarization

Since starting with the era when first-time automatic text summarization came into the picture (Luhn, 1958), the text processing task is performed mainly by using features based on IR (Information Retrieval) measures, i.e., term frequency (TF), inverse term frequency (TF-IDF). Table-1 shows a detailed survey on extractive text summarization with research papers, the dataset used, the system's accuracy, and its pros and cons.Earlier, the efficiency of the summary was prepared by the proportion of no. of judged-important points to total no. of words in the summary (Garner, 1982). The immediate summarization result and relationship to detailed comprehension and recall results were analyzed in that study. The lack of linguistic knowledge is a weak point for extracting helpful information from a large amount of the data. To overcome these two limitations: (i) One mechanism that deals with unknown words and gaps in linguistic information. (ii) To extract linguistic information from text automatically, SCISOR (System for Conceptual Information Summarization, Organization and Retrieval) was developed by Rau et al. (1989). Experiments performed on summarization until 1990 were focused on just extracting (reproduction) the summaries from original text rather than abstracting (newly generated). SUMMRIST system (Hovy & Lin, 1996) was developed with the help of NLP techniques, in which one can create a multi-lingual summarizer by modifying some part of the structure.

The challenges with traditional frequency-based, knowledge-based and discourse-based summarization lead to encountering these challenges with robust NLP techniques like corpus-based statistical NLP (Aone et al., 1997). The summarization system named DimSum consists of a summarization server and summarization client. The features produced from these powerful NLP algorithms were also used to give the user numerous summary views in an innovative way. Evaluating the summaries of humans and systems by four parameters; optimistic evaluation, pessimistic evaluation, intersection evaluation, union evaluation(Salton et al., 1997) and proven that the summaries generated by the two humans are dissimilar for the same article while automatic methods are favourable here. A robust summarization was practically implemented on online news 'New York Times' by Tomek (1998), which gives summaries very quickly with including significantly less portion of original lengthy text. The study of effects of headings on text summarization proven (Lorch et al., 2001) that readers depend heavily on organizational signals to construct a topic structure. The machine learning approach (Neto et al., 2002) considers automatic text summarization as a two-class classification problem, where a sentence is considered 'correct' if it appears in extractive reference summary or otherwise as 'incorrect'. Here they used two famous ML classification approaches, Naïve Bayes and C4.5. Lexicon is a salient part of ant textual data. Focusing on an algorithm (Silber & McCoy, 2002) that efficiently makes lexical chains in linear time is a feasible intermediate representation of text summarization.

As a prior study shows, supervised methods were where human-made summaries helped us find parameters or features of summarization algorithms. Despite that, unsupervised methods (Nomoto & Matsumoto, 2003) with diversity functionality define relevant features without any help from human-made summaries. (Yeh et al., 2005b) proposed a trainable summarizer that generates summaries based on numerous factors such as location, positive keyword, negative keyword, centrality, and resemblance to the title. It uses a genetic algorithm (GA) to train the score function to find a good combination of feature weights. After that, it employs latent semantic analysis (LSA) to derive a document's or corpus' semantic matrix and semantic sentence representation to build a semantic text relationship map. Combining three approaches: a diversity-based method, fuzzy logic, and swarm-based methods (Binwahlan et al., 2010), can generate good summaries. Where diversity-based methods use to figure out similar sentences and get the most diverse sentence and concentrate on reducing redundancy, while swarm-based methods are used to distinguish the most important and less important sentences then use fuzzy logic to tolerate redundancy, approximate values and uncertainty, and this combination concentrates on the scoring techniques of sentences. While comparing two-approach Swarm-fuzzy based methods performs well than diversity-based methods here.

The construction of methods for measuring the efficiency of SAS (Systems of automatic summarization) functioning is an important area in the theory and practice of automatic summarization. Based on a model vocabulary supplied by subjects, four techniques (ESSENCE (ESS), Subject Search Summarizer (SSS), COPERNIC (COP), Open Text Summarizer (OTS)) of automatic text summarization are evaluated by Yatsko & Vishnyakov (2007). The distribution of vocabulary terms in the source text is compared to the distribution of vocabulary terms in summaries of various lengths generated by the systems. (Ye et al. (2007b) contend that the quality of a summary can be judged by how many concepts from the source documents can be retained after summarization. As a result, summary generation can be viewed as an optimization task involving selecting a set of sentences with the least answer loss. The proposed document concept lattice (DCL) is a unique document model that indexes sentences based on their coverage of overlapping concepts. The authors of (Ko & Seo, 2008) suggested method merges two consecutive sentences into a bi-gram pseudo sentence, allowing statistical sentence-extraction tools to use contextual information. The statistical sentence-extraction approaches first choose salient bi-gram pseudo sentences, and then each selected bi-gram pseudo sentence is split into two single sentences. The second sentence-extraction operation for the separated single sentences is completed to create a final text summary.

CN-Summ(Complex Networks-based Summarization) was proposed by (Antiqueira et al., 2009). Nodes relate to sentences in the graph or network representing one piece of text, while edges connect sentences that share common significant nouns. CN-Summ consists of 4 steps: 1) prepossessing (lemmatization). 2) resulting text is mapped to a network representation according to adjutancy and weight metrics of order n*n (n is no. of sentences/nodes) .3) compute different network measurements 4) the first m sentences are selected as summary sentences depending upon compression rate. Alguliev & Aliguliyev (2009) gave a new approach for unsupervised text summarization. That approach is focused on sentence clustering, where clustering is the technique of detecting interesting distributions and patterns within multidimensional data by establishing natural groupings or clusters based on some similarity metric. Here the researchers have proposed a new method to measure similarity named Normalized Google Distance (NGD) and to optimize criterion functions discrete differential evolution algorithm called as MDDE (Modified Discrete Differential Evolution) Algorithm is proposed.

Swarm Intelligence (SI) is the collective intelligence resulting from the collective behaviours of (unsophisticated) individuals interacting locally and with their environment, causing coherent functional global patterns to emerge. The primary computational parts of swarm intelligence are Particle Swarm Optimization (PSO), which is inspired by the social behaviour of bird flocking or fish schooling, and Ant Colony Optimization (ACO), which is inspired by the behaviour of ants. Binwahlan et al. (2009a) suggested a model based on PSO whose primary goal is to score sentences while focusing on dealing equally with text elements depending on their value. Combining three approaches: diversity-based method, fuzzy logic, and swarm-based methods (Binwahlan et al., 2010) can generate good summaries. Where diversity-based methods use to figure out similar sentences and get the most diverse sentence and concentrate on reducing redundancy, while swarm-based methods are used to distinguish the most important and less important sentences then use fuzzy logic to tolerate redundancy, imprecise values and uncertainty, and this combination concentrates on the scoring techniques of sentences. While comparing two-approach Swarm-fuzzy based methods performs well than diversity-based methods here.

In (Mashechkin et al., 2011), the researchers had used LSA(Latent Semantic Analysis) for text summarization. The original text is reproduced as a matrix of terms and sentences. Text sentences are represented as vectors in the term space, and a matrix column represents each sentence. The resultant matrix is then subjected to latent semantic analysis to construct a representation of text sentences in the topic space, which is performed by applying one of the matrix factorizations (singular value decomposition (SVD)) to the text matrix. (Alguliev et al., 2011b) consider text summarization problem as integer linear programming problem while assuming that summarization is a task of finding a subset of sentences from the original text to represent important detail of the original text. That study focused on three characteristics (relevance, redundancy, and length) and tried to optimize that by particle swarm optimization algorithm (PSO) and branch and bound optimization algorithm. In extractive text summarization, sentence scoring is the most commonly used technique. The study (Ferreira et al., 2013) evaluated 15 algorithms available for sentence scoring based on quantitative and qualitative assessments. In conjunction with a graph-based ranking summarizer, Wikipedia is given by (Sankarasubramaniam et al., 2014a). It has given a unique concept by introducing incremental summarization property, where single and multi-document both can provide additional content in real-time. So, the users can first see the initial summary, and if willing to see other content, they can make a request.

Using a deep auto-encoder (AE) to calculate a feature space from the term-frequency (tf) input, (Yousefi-Azar & Hamey, 2017b) offer approaches for extractive query-oriented single-document summarization. Both local and global vocabularies are considered in experiments. The study shows the effect of adding slight random noise to local TF as the AE's input representation and propose the Ensemble Noisy Auto-Encoder as a collection of such noisy AEs (ENAE). Even though there is a lot of study on domain-based summarising in English and other languages, there is not much in Arabic due to a lack of knowledge bases. A hybrid, single-document text summarization approach is proposed in (Al-Radaideh & Bataineh, 2018a) paper (ASDKGA). The method uses domain expertise, statistical traits, and genetic algorithms to extract essential points from Arabic political documents. For domain or genre-specific summarization (such as for medical reports or specific news articles), feature engineering-based models have shown to be far more successful, as classifiers can be taught to recognize particular forms of information. For general text summary, these algorithms produce poor results. To overcome the issue, an entirely data-driven approach for automatic text summarization is given by (Sinha et al., 2018)

The most challenging difficulties are covering a wide range of topics and providing diversity in summary. New research based on clustering, optimization, and evolutionary algorithms has yielded promising results for text summarization. A two-stage sentences selection model based on clustering and optimization techniques, called COSUM, was proposed by Alguliyev et al. (2019b). The sentence set is clustered using the k - means algorithm in the first stage to discover all subjects in a text. An optimization approach is proposed in the second step for selecting significant sentences from clusters. The most crucial reason for the lack of domain shift approaches could be understanding different domain definitions in text summarization. For the text summarization task, (Wang et al., 2019) extended the traditional definition of the domain from categories to data sources. Then used, a multi-domain summary dataset to see how the distance between different domains affects neural summarization model performance. Traditional applications have a major flaw: they use high-dimensional, sparse data, making it impossible to gather relevant information. Word embedding is a neural network technique that produces a considerably smaller word representation than the classic Bag-of-Words (BOW) method. (Alami et al., 2019) has created a text summarization system based on word embeddings, and it showed that the Word2Vec representation outperforms the classic BOW representation. Another summarization approach using word embeddings was given by (Mohd et al., 2020). This study also used Word2Vec as a distributional semantic model that captures the semantics.

Current state-of-art systems produce generic summaries that are unrelated to the preferences and expectations of their users. CTRLsum (He, Kryscinski, et al., 2020), a unique framework for controlled summarizing, is presented to address that limitation. This system permits users to interact with the summary system via textual input in a collection of key phrases or descriptive prompts to influence several features of generated summaries. The majority of recent neural network summarization algorithms are either selection-based extraction or generation-based abstraction. (Xu & Durrett, 2020) introduced a neural model based on joint extraction and syntactic compression for single-document summarization. The model selects phrases from the document, identifies plausible compressions based on constituent parses, and rates those compressions using a neural model to construct the final summary. Four algorithms were proposed by (El-Kassas et al., 2020). The first algorithm uses the input document to create a new text graph model representation. The second and third algorithms look for sentences to include in the candidate summary in the built text graph. The fourth algorithm selects the most important sentences when the resulting candidate summary exceeds a user-specified limit. Automatic text summarization is an arduous effort for under-resourced languages like Hindi, and it is still an unsolved topic. Another problem with such languages is the lack of corpus and insufficient processing tools. For Hindi novels and stories, (Rani & Lobiyal, 2021) developed an extractive lexical knowledge-rich topic modelling text summarising approach in this study. The standard words-based similarity measure grants weight to most graph-based text summarising techniques. Belwal et al. (2021) offered a new graph-based summarization technique that considers the similarity between individual words and the sentences and the entire input text.

Table 1: Research survey on Extractive text summarization method

| Citation | article | Model/methods/techniques applied | Dataset Used | Performance | Advantages/Pros | Disadvantages/Cons |
|---|---|---|---|---|---|---|
| (Garner, 1982) | Efficient Text Summarization: Costs and Benefits | Extractive text summarization | Dutch elm Disease (167 words- a single word) | Efficiency: proportion of number of judged-important ideas to total number of words in summary Proportion=0.2 to 0.12 | - | - |
| (Rau et al., 1989) | Information extraction and text summarization using linguistic knowledge acquisition | Extractive text summarization | Dataset named as: Group Offers to Sweeten Warnaco Bid | - | SCISOR is robust and reliable extraction of information | Size of lexicon. (10,000 words) and system is limited. |
| (Bloom et al., 1994) | Automatic Analysis, Theme Generation, and Summarization of Machine-Readable Texts | Extractive text summarization | 175-page article Entitle "United States of America. | - | robust and generally applicable to a wide variety of texts in many different environments, | In the absence of deep linguistic, it is not possible to build Intellectually satisfactory text summaries. |
| (Reimer et al., 1997) | Formal Model of Text Summarization Based on Condensation Operators of a Terminological Logic | Extractive text summarization | text | - | attempt was made to properly integrate the text summarization process to the formal reasoning mechanisms of the underlying knowledge representation language | Currently the summarization process considers only activity and connectivity patterns m the text knowledge base |
| (Aone et al., 1997) | A Scalable Summarization System Using Robust NLP | Extractive text summarization | text | - | The DimSum summarization system advances summarization technology by applying corpus-based statistical NLP techniques, robust information extraction, and readily available on-line resources | The results are not evaluated with human generate summaries. |
| (Salton et al., 1997) | Automatic text structuring and summarization | Multi document extractive text summation | encyclopaedia article 78 (Abortion) | 4 factors are compared with respect to human extract: Optimistic evaluation, Pessimistic evaluation, Intersection, and Union. | Original text is broken into constituent pieces and combing these pieces according to their functionality. This text structure used for producing comprehensive summaries by automatic paragraph extraction -domain independent | Human extracts are dissimilar for a same article |
| (Lorch et al., 2001) | Effects of Headings on Text Summarization | Extractive text summarization | The experimental text was a revised version of a text on energy problems and solutions | - | More topics were included in the summaries of participants who read the text with headings than in the summaries of participants who read the text without headings | Headings interacted in recall, but not in the summarization task |
| (Neto et al., 2002) | Automatic Text Summarization Using a Machine Learning Approach | Extractive summarization on multi-document by machine learning approach | TIPSTER document base. | Comparison between automatically produced and manually produced summaries: Compression rate: 10% Precision/Recall: 30.79 ± 3.96 | Works well with Naïve Bayes although the dataset consist of 33,658 documents within | C4.5 gives poor results as compared to naïve Bayes. |
| (Silber & McCoy, 2002) | Efficiently computed lexical chains as an intermediate representation for automatic text summarization | Multi document extractive text summarization | Randomly selected documents which are of length ranging from 2,247 to 26,320 words each. | Time complexity of algorithm: linear time (O(n) where n is nouns in document) | Feasible and robust algorithm with linear time complexity | Results are poor where proper nouns as proper nouns and domain specific are not found in WordNet. |
| (Nomoto & Matsumoto, 2003) | The diversity-based approach to open-domain text summarization | Unsupervised text summarization and open domain | 5080 news articles in Japanese | Information centric evaluation | -Does not rely on human-made summaries but measuring the performance in terms of IR. -Diversity based summarizer is superior to tf-idf based summaries. | Diversity based summarizer perform poorly when exhibit high agreement. |
| (Yeh et al., 2005a) | Text summarization using a trainable summarizer and latent semantic analysis | modified corpus-based approach (MCBA) and LSA-based T.R.M. approach | 100 documents in 5 sets of domains of politics were collected from New Taiwan Weekly | Compression rate: 30% MCBA: F-measure: 49% MCBA+GA: F-measure: 52% LSA+T.R.M F-measure: 44% (for single-document) F-measure: 40% (for corpus) | -LSA+T.R.M. performs better than keyword-based text summarization. -The results of MCBA+GA show that using GA in training is a good technique to deal with the circumstance where we are undecided about an appropriate combination of feature weights. | The performance of both approaches vary as compression rates are changes. For good result chose appropriate compression rate. |
| (Ye et al., 2007a) | Document concept lattice for text understanding and summarization | multi-document extractive summarization | DUC-2005 DUC-2006 | -ROUGE-2 Recall = 7.17% & ROUGE-SU4 Recall = 13.16% in DUC-2005 -ROUGE-2 Recall = 8.99% and ROUGE-SU4 | Proposed a document concept lattice (DCL) model and the corresponding algorithm for summarization | - |
| | | | | Recall = 14.75% in DUC-2006 | | |
| (Ko & Seo, 2008) | An effective sentence-extraction technique using contextual information and statistical approaches for text summarization | Single/multi-document extractive summarization | 841 news articles from KOrea Research and Development Information Centre (KORDIC) dataset | F1 score: 47.9 with 10% summary<br>F1 score: 50.4 with 30% | To tackle the feature sparseness problem, it used contextual information (bi-gram pseudo phrase) and the combination method of statistical methodologies to increase the performance.<br>-Independent on language | The performance changes according to no. of query words in multi-document |
| (Antiqueira et al., 2009) | A complex network approach to text summarization | graph-based extractive summarization | 100 newspaper articles in Brazilian Portuguese | Average f-measure for all 15 version: 42% | Language & domain independent graph-based approach using complex networks (CN-Summ) | There are 15 versions of CN-Summ. Not an approach that combine all. |
| (alguliev & aliguliyev, 2009) | Evolutionary Algorithm for Extractive Text Summarization | Unsupervised extractive text summarization | DUC2001 & DUC2002 | ROUGE-1: 0.45952<br>ROUGE-2: 0.19338<br>ROUGE-L: 0.21763 | Proposed new method for measuring similarity between sentences<br>- proposed a new algorithm method to optimize objective functions | -the results is shown using 3 criterion functions and among them criterion function 1 perform poorly. |
| (Binwahlan et al., 2009b) | Swarm Based Text Summarization | Swarm intelligence based extractive text summarization | DUC2002 | The proposed model creates summaries which are 43% similar to the manually generated summaries | Employed PSO for sentence scoring based on their importance. | The experiment results are compared with manually human made summaries which are only 49% similar. |
| (Binwahlan et al., 2010) | Fuzzy swarm diversity hybrid model for text summarization | Extractive text summarization (Diversity-swarm-fuzzy) | First 100 documents from DUC 2002. | ROUGE-1:<br>Fuzzy swarm<br>Average F: 0.45524 | Diversity-based and fuzzy methods reduce ambiguity.<br>Swarm-based methods differentiate important sentences and less important sentences | Well accepted method as per results but not made for multi-document summarization. |
| (Mashechkin et al., 2011) | Automatic text summarization using latent semantic analysis | LSA based extractive text summarization. | DUC2001 & DUC2002 | ROUGE-2: 0.19260<br>ROUGE-L: 0.37229 | A new generic summarization method is proposed that uses nonnegative matrix factorization to estimate sentence relevance<br>-In comparison to the singular value decomposition position, the proposed estimation of semantic characteristics (topics) weight better preserves the internal structure of the text. | The optimal numbers of topics for each summarization problem is empirically obtained. |
| (Alguliev et al., 2011a) | MCMR: Maximum coverage and minimum redundant text summarization model | PSO, unsupervised based multi-document extractive text summarization | DUC2005 & DUC2007 | ROUGE-2 on DUC-2005:<br>B&B: 0.0790<br>PSO: 0.0754<br>ROUGE-2 on DUC-2007:<br>B&B: 0.1221<br>PSO: 0.1165 | Good for both single document and multi-document text summarization | -Results are directly depending upon optimization algorithm<br>-parameter value of objective function influences performance of summarization |
| (Sankarasubramaniam et al., 2014b) | Text summarization using Wikipedia | Wikipedia-based graph-based multi-document summarization algorithm | DUC 2002 | ROUGE-1: 0.46<br>ROUGE-2: 0.23<br>Precision: 0.57<br>Recall:0.50<br>F-measure: 0.51 | To extract salient topics from text document it proposed algorithm using Wikipedia with graph-based summarization<br>-concentrated on personalization and query-focusing of summaries | Among results of evaluation Sentences chosen uniformly at random performs poor. |
| (Gupta & Kaur, 2016) | A Novel Hybrid Text Summarization System for Punjabi Text | Extractive text summarization (concept, statistical, location, numeric data, and linguistic-based features) | Standard Unicode-based Punjabi Corpus and Punjabi news corpus AZIT with each having 150 documents. | Overall score ROUGE-2: 0.85<br>F-score with 50% compression rate: 0.86 | Question-answering summary also tried here where proposed method gives 80-90% accuracy. | Language dependent |
| (Yousefi-Azar & Hamey, 2017a) | Text summarization using unsupervised deep learning | query-oriented single-document extractive text summarization using deep learning | Summarization and Keyword Extraction from Emails (SKE)<br>BC3 from British Columbia University | Improvement of ROUGE-2 recall on average 11.2%. | Auto-Encoders produces a rich concept vector for an entire sentence from a bag-of-words input. | -Better to use semi-supervised technique because shortage of manually annotated data<br>-the computational cost of training and the necessity to appropriately tune the training hyper parameters |
| (Sinha et al., n.d.) | Extractive Text Summarization using Neural Networks | Extractive text summarization using Neural Networks | DUC 2002 | ROUGE-1: 55.1<br>ROUGE-2: 22.6 | A data-driven technique that leverages a basic feedforward neural network achieves good performance while compared with state-of-the-art systems while being both implementationally and computationally light. | assumed that summary length to be generated should be less than specified length. |
| (Al-Radaideh & Bataineh, 2018b) | A Hybrid Approach for Arabic Text Summarization Using Domain Knowledge and Genetic Algorithms | Extractive text summarization (domain knowledge, genetic algorithm) | KALIMAT corpus and Essex Arabic Summaries Corpus (EASC) | Average F-measure: 0.605 with compression ratio of 40%. | Domain knowledge enhances results of the proposed method | -Language dependent<br>-training done with only 500 keywords which lies into 4 categories. |
| (Alguliyev et al., 2019a) | COSUM: Text summarization based on clustering and optimization | Extractive text summarization using clustering and optimization | DUC2001 & DUC2002 | DUC2001:<br>ROUGE-1: 0.4727<br>ROUGE-2: 0.2012<br><br>DUC2002:<br>ROUGE-1: 0.4908<br>ROUGE-2: 0.2309 | The model additionally limits the length of sentences selected in the candidate summary to ensure readability. | Sometimes performance is conceded while compared to other methods |
| (Wang et al., 2019) | Exploring Domain Shift in Extractive Text Summarization | Extractive multi-domain text summarization | MULTI-SUM dataset<br>CNN/DailyMail | - | Proposed a multi-domain dataset MULTI-SUM<br>-Given four learning schemes as a preliminary investigation into the properties of various learning strategies when dealing with multi-domain summarising jobs. | Excellent results in cross-dataset (CNN/DailyMail) but results declined in in-domain and out-of-domain settings. |
| (Xu & Durrett, 2020) | Neural Extractive Text Summarization with Syntactic Compression | Extractive text summarization with deep neural network | New York Times corpus<br>CNN/DailyMail | CNN:<br>ROUGE-1: 32.7<br>ROUGE-2: 12.2<br>ROUGE-L: 29.0 | A sentence extraction model is combined with a compression classifier to determine whether or not syntax-derived compression choices should be deleted for each sentence. | Best performance on CNN but performance somewhere declined for another dataset while compared to baseline methods. |
| (Mohd et al., 2020) | Text document summarization using word embedding | Extractive text summarization using distributional semantics (Word2vec) | DUC2007 | F-score for 25% summary length:<br>ROUGE-1: 33%<br>ROUGE-2: 7%<br>ROUGE-L: 20%<br>ROUGE-SU4: 13% | summarization technique based on the distributional hypothesis to capture the semantics of better than the baselines. | Proposed distributed semantic model is computationally expensive and time consuming.<br>Results are given on different % of summary length, where sometimes recall values are low than baseline systems. |
| (He, Kryściński, et al., 2020) | CTRLsum: Towards Generic Controllable Text Summarization | Extractive text summarization (keyword-based model) | CNN/DailyMail arXiv scientific papers BIGPATENT patent articles | CNN/DailyMail: ROUGE-1: 48.75 ROUGE-2: 25.98 ROUGE-L: 45.42 arXiv: ROUGE-1: 47.58 ROUGE-2: 18.33 ROUGE-L: 42.79 Success rate: 97.6 Factual Correctness: 99.0 | proposed a generic framework to perform multi-aspect controllable summarization. | During training, the model is conditioned on keywords to predict summaries but one keyword can be fall into more than one aspect. |
| (El-Kassas et al., 2020) | EdgeSumm: Graph-based framework for automatic text summarization | Extractive single-document text summarization using Graph-based approach | DUC2001 & DUC2002 | Improvement of 1.2% and 4.7% in ROUGE-1 and ROUGE-L respectively for DUC2002 dataset. | -EdgeSumm integrates a number of extractive ATS methods (graph-based, statistical-based, semantic-based, and centrality-based methods) to take use of their strengths while minimising their weaknesses.<br>-The use of hyphens removal and synonyms substitution tasks on the performance of the generated summaries is investigated. | -Results are very promising but not extended for multi-document summarization<br>-not domain specific system<br>-Language dependent |
| (Rani & Lobiyal, 2021) | An extractive text summarization approach using tagged-LDA based topic modelling | Extractive text summarization based on LDA-model | 114 Hindi novels including short stories from 'Munshi Premchand's stories' blog | Perform preferable than the baseline algorithms for 10% - 30% compression ratios and given evaluation metrics | -Since there was no corpus of Hindi novels and stories, it built one corpus.<br>-four distinct sentence weighting scheme-based variants are derived by manipulating the proposed system | -Semantic features are not taken into consideration<br>-Results are somewhere improved and somewhere declined while compared to baseline systems for different-different compression rates |
| (Belwal et al., 2021) | A new graph-based extractive text summarization using keywords or topic modelling | Extractive summarization using Graph-based and Topic-based techniques | Opinosis dataset CNN/ DailyMail | CNN/DailyMail: ROUGE-1: 0.428 ROUGE-2: 0.201 ROUGE-L: 0.392<br><br>Opinosis dataset: ROUGE-1: 0.271 ROUGE-2: 0.084 ROUGE-L: 0.161 | Added an extra parameter that calculates the similarity of the nodes to the entire content.<br>Used topic modelling to address the issue of redundancy associated with existing summarising methods | - |
| (Alami, Mallahi, et al., 2021) | Hybrid method for text summarization based on statistical and semantic treatment | Extractive text summarization (statistical and semantic methods) | The dataset consists of 153 Arabic articles taken from two Arabic newspapers and the Arabic version of Wikipedia. The | Dataset-1: F1-measure: 59.47 with 40% compression rate Dataset-2: F1-measure:60.79 40% compression rate | Remove redundancy with using MMR (Maximal Marginal Relevance)<br>Improve sentence score by other statistical features | Language dependent |

## 3.2 Abstractive text summarization

The competitions DUC-22003 and DUC-2004 had standardized the task of Abstractive text summarization, in which news articles from various fields with multiple reference summaries per article generated by humans are used as datasets. The TOPIARY system (Zajic et al., 2004) stood the best performing technique. Some noticeable work was submitted by Banko et al. (2000) with phrase-table based machine translation techniques and (Woodsend et al., 2010) with quasi-synchronous grammar techniques. Table-2 shows a detailed survey on Abstractive text summarization with a particular research paper with the dataset used, the system's accuracy, and its pros and cons.After that, deep learning was introduced as a viable alternative to many NLP problems. Text is a sequence of words where sequence-to-sequence models can entertain input and output sequences. With the apparent similarities Machine translation (MT) problem may be mapped to text summarization despite that abstractive summarization is very different from it. MT is lossless while summarization is lossy in the manner, and MT is a one-to-one word-level mapping between source and target, but that mapping is less in summarization.

In (Rush et al., 2015), the researchers had used convolution models to encode the input and context-sensitive feed-forward network with attentional mechanism and showed better results for Gigaword and DUC datasets. (Chen, 2015) have produced a sizeable Chinese dataset for short text summarization (LCSTS), which has given good results on their dataset while using RNN architecture at both encoder and decoder sides. Beyond RNN architecture at both encoder and decoder sides, (Nallapati et al., 2016) captured keywords, modelled unseen or rare words, and captured the document's hierarchy using a hierarchical attention mechanism. The authors have also tried to analyse the quality of the output summary. In that case, somewhere models perform well and somewhere poor compared to others. Human's summaries are more abstractive naturally because they use some inherent structures while writing summaries. The deterministic transformation in a discriminative model(RNN) used by Nallapati et al. (2016) limits the representation of latent structure information. After that, Miao & Blunsom (2016) gave a generative model to capture the latent structure, but they did not consider recurrent dependencies in their generative model. The authors of (Li et al., 2017) tried to find some common structures such as "what", "what happened", "who actioned what" from the source and proposed a deep recurrent generative model for modelling latent structure.

AMR (Abstract Meaning Representation) was firstly introduced by Banarescu et al. (2013). AMR targets fetching the meaning of the text by giving a special-meaning representation to the source text. AMR attempts to capture "who is doing what to whom". The work of Liu et al. (2015).' s includes AMR, but they did not use it at the abstraction level, so their work is limited to extractive summarization only. Also, the approach aims to generate a summary from a story. Producing a single graph assumes that all the important sentences can be extracted from a single subgraph. Difficulties arise when information is spread out all over. So, Doha et al. (2017a) worked on multiple summary graphs and explored problems with existing evaluation methods and datasets while doing abstractive summarization.

Combining the advantages of extractive and abstractive and curing the disadvantages (Song et al., 2019b) had implemented a model named ATSDL (ATS using DL). This model uses a phrase extraction method called MOSP to extract key phrases from the original text after that, learns the collocation of phrases. Following training, the model will generate a phrase sequence that satisfies the syntactic structure criteria. Furthermore, we leverage phrase location information to overcome the problem of unusual terms, which practically all abstractive models would face. Regarding sequence-to-sequence models, RNN is not most used because it tends to low-efficiency problems as they rely on the previous step when training, and it must preserve the hidden state of the whole past, thus not able to perform parallel operations. To overcome these problems, Zhang et al. (2019) proposed a sequence-to-sequence model based on CNN to create the representation of source text. As we know, traditional CNN can only encode with fixed size contexts of inputs, but in this study, they increase the compelling text by stacking CNN layers over each other. The length of the sequence under consideration may thus be readily regulated, and each component of the sequence can be computed in parallel. More commonly, abstractive summarization problems are that the generated summaries are frequently incompatible with the source content in terms of semantics. WEI et al. (2018) offer a regularisation strategy for the sequence-to-sequence model in this research, and we use what the model has learnt to regularise the learning objective to mitigate this problem's influence.

Until now, the model discussed does not consider whether the summaries are factually consistent with source documents. Kryściński et al. (2019a) present a model-based technique for evaluating factual consistency and detecting conflicts between source documents and the output summary that is a weakly supervised model. The steps of these models are:

- Determine whether sentences are factually consistent after being transformed,
- Find a span in the source documents to validate the consistency prediction, and
- Find an inconsistent span in the summary phrase if one exists.

Another notable work is done in the sequence-to-sequence encoder and decoder approach by Kryściński et al. (2020). That study makes two main contributions. First, separate extraction and generation at decoder part. The contextual network is standalone for extraction, and a language model generates paraphrases. Second, optimizing the n-gram overlap while encouraging abstraction with ground-truth summaries.

Wang et al. (2020) provide a unique Generative Adversarial Network (GAN) for Abstractive Text Summarization with a multitask constraint in this research (PGAN-ATSMT). Through adversarial learning, this model simultaneously trains a generator G and a discriminator D. The sequence-to-sequence architecture is the backbone of the generative model G, which takes the source document as input and generates the summary. The model uses a language model to implement D instead of a binary classifier as the discriminative model D, and the output of the language model is used as the reward to steer the generative model. A minimax two-player game was used to optimize the generative model G and the discriminative model D. Extended work on the GAN network is done by Yang et al. (2021). They present a new Hierarchical Human-like deep neural network for ATS (HH-ATS), influenced by how humans interpret articles and produce summaries. HH-ATS comprises three main components (i.e., a knowledge-aware hierarchical attention module, a multitask learning module, and a dual discriminator generative adversarial network) that reflect the three phases of human reading cognition (i.e., rough reading, active reading, and post-editing).

Table 2: Research survey on Abstractive text summarization method

| Citation | article | Model/methods/techniques applied | Dataset Used | Performance | Advantages/Pros | Disadvantages/Cons |
|---|---|---|---|---|---|---|
| (Nallapati et al., 2016) | Abstractive text summarization using sequence-to-sequence RNNs and beyond | Abstractive (Sequence-to-sequence encoder-decoder with RNN) | Gigaword corpus, DUC, CNN/daily mail corpus | ROUGE-1: 35.46<br>ROUGE-2:13.30<br>ROUGE-L:32.65<br>Qualitative evaluation: a few high quality and poor-quality output | -Performs a little bit well as compared to state-of-the-art<br>-Propose new dataset for multiple sentence summary | -Worked for single sentence output summaries<br>-capturing the meaning of complex sentence is weakness for this model |
| (Li et al., 2017) | Deep recurrent generative decoder for abstractive text summarization | Abstractive (Sequence-to-sequence encoder-decoder with deep recurrent generative decoder (DRGN)) | Gigaword corpus, DUC, LCSTS | ROUGE-1: 36.71<br>ROUGE-2:24.00<br>ROUGE-L:34.10 | Implement deep recurrent generative model to capture latent structure information. | Worked for single sentence output summaries |
| (Dohare et al., 2017b) | Text Summarization using Abstract Meaning Representation | Abstractive summarization with AMR. | CNN, DailyMail | ROUGE-1: 39.53<br>ROUGE-2:17.28<br>ROUGE-L:36.38 | -Suggested a full pipeline for summarization with AMR<br>-proven that ROGUE can't be used for evaluating the abstractive summaries<br>-a novel approach for extracting multiple summary graphs | Not a lot of work has been done to extract AMR graphs for summaries |
| (Song et al., 2019a) | Abstractive text summarization using LSTM-CNN based deep learning | Abstractive (CNN-LSTM) | CNN, DailyMail dataset | ROUGE-1: 34.9<br>ROUGE-2:17.8 | -Implement ATSDL (ATS using Deep Learning) system that Combines CNN and LSTM for better performance<br>-solve problem of rare words | -Training of deep learning is very time-consuming process.<br>-ROUGE can't evaluate the quality of summary effectively. |
| (Zhang et al., 2019) | Abstract text summarization with a convolutional seq2seq model | Abstractive (CNN) | Gigaword corpus, DUC, CNN/daily mail corpus | ROUGE-1: 42.04<br>ROUGE-2: 19.77<br>ROUGE-L: 39.42 | -equip the CNN model with GLU and residual connections.<br>-hierarchical attention mechanism to generate the keywords and the key sentences simultaneously.<br>-a copying mechanism to extract out-of-vocabulary words from source text. | In practise, adding more sentences has a negative impact on performance, which we explain to the fact that the latter sentences are unrelated to the summary. |
| (Kryściński et al., 2019b) | Evaluating the Factual Consistency of Abstractive Text Summarization | Abstractive | CNN/Daily Mail dataset | Accuracy: 74.15<br>F1-Score: 0.5106 | -Implemented factual consistency checking model (FactCC). | -common-sense mistakes made by summarization models.<br>- stemming errors from dependencies between different sentences within the summary |
| (WEI et al., 2018) | Regularizing output distribution of abstractive Chinese social media text summarization for improved semantic consistency | Abstractive | Large-Scale Chinese Short Text Summarization Dataset (LCSTS) | ROUGE-1: 36.2<br>ROUGE-2:24.3<br>ROUGE-L: 33.8<br>Accuracy of Human Evaluation approach: 53.6% | - suggest a method for regularising the output word distribution so that semantic inconsistency in the training data, such as terms not linked to the source content, is underrepresented in the model.<br>-Proposed a simple human evaluation approach for determining the generated summary's semantic compatibility with the original information. | Language dependent (Chinese dataset) |
| (Kryściński et al., 2018) | Improving abstraction in text summarization | Abstractive (Sequence-to-sequence encoder-decoder) | CNN/Daily Mail dataset | ROUGE-1: 40.19<br>ROUGE-2:17.38<br>ROUGE-L: 37.52<br>Qualitative evaluation:<br>Readability: 6.76/10<br>Relevance: 6.73/10 | -Separate the decoder into a contextual network and a pretrained language model.<br>-Optimized metric through policy learning | Matches score when compared to the state-of-the-art in relevance but slightly low in terms of readability of summaries. |
| (Yang, Wang, et al., 2020) | Plausibility-promoting generative adversarial network for abstractive text summarization with multi-task constraint | Abstractive (GAN) | Gigaword corpus, CNN/daily mail corpus | ROUGE-1: 40.19<br>ROUGE-2:17.38<br>ROUGE-L: 37.52<br>Perplexity:10.21<br>Qualitative evaluation:<br>Relevance: 3.29/5<br>Fluency:3.38/5 | -Developed PGAN-ATSMT, an adversarial framework for abstractive text summarization with multi-task constraint.<br>-PGAN-ATSMT jointly trains the task of abstractive text summarization and two other related tasks: text classification and syntax generation. | -External Common-sense knowledge from the language is missing.<br>-Time consuming as deep learning algorithm |
| (Yang, Li, et al., 2020) | Hierarchical Human-Like Deep Neural Networks for Abstractive Text Summarization | Abstractive (GAN) | Gigaword corpus, CNN/daily mail corpus | ROUGE-1: 43.16<br>ROUGE-2:20.32<br>ROUGE-L: 39.14<br>Qualitative evaluation:<br>Informativeness: 3.41/5<br>Fluency:3.32/5 | HH-ATS extends the seq2seq framework by replicating the process of writing a summary for a piece of text by humans. | To evaluate human judgement better it is required to explore different automatic metrics |

### 3.3 Hybrid text summarization:

When discussing very precious approaches of Automatic text summarization that are Extractive and Abstractive, both come with their pros and cons. Extractive summarization is comparatively easier to implement than abstractive summarization, but extractive summarization is not as efficient as user perception. Combining these methods by strengthening their pros and weakening their cons leads to hybrid methods for text summarization.

Experiments were done on summarization until 1990 were focused on just extracting (reproduced) the summaries from original text rather than abstracting (newly generated).SUMMRIST system (Hovy & Lin, 1996) was developed with the help of NLP techniques. We can develop a multi-lingual summarizer by modifying some parts of the structure.

Semantic and statistical features combine extracting and abstracting. The authors of Bhat et al. (2018) used emotions of the text as a semantic feature. Emotions play a significant role in defining the user's emotional affinity, so lines with implicit emotional content are crucial to the writer and should be included in the summary. The extracted summary is then put into the Novel language generator, a hybrid summarizer that combines WordNet, Lesk algorithm, and POS to transform extractive summary into an abstractive summary. Table-3 shows a detailed survey on hybrid text summarization with a particular research paper with the dataset used, accuracy of the system and its pros and cons.

Table 3. Research survey on hybrid text summarization method

| Citation | article | Model/methods/techniques applied | Dataset Used | Performance | Advantages/Pros | Disadvantages/Cons |
|---|---|---|---|---|---|---|
| (Hovy & Lin, n.d.) | Automated text summarization and the SUMMARIST system | Hybrid text summarization | Text | SUMMARIST scores 30% higher than human abstracts | By avoiding some language-specific methods in SUMMARIST it is possible to create multi-lingual summarizer easily. | Lack of good training data in variety of domains |
| (Bhat et al., 2018) | SumItUp: A Hybrid Single-Document Text Summarizer | Hybrid Single-Document Text summarization | DUC-2007 with 15 news articles | Precision:0.4 to 0.8<br>Compression rate: 35% | -hybrid approach to a single-document summarization combining semantic and statistical features<br>-emotional feature | Not proper evaluation is done. The results are compared to MS word (no ROUGE metrics) |

## 4 ATS System Evaluation and Evaluation Programs

There have been many efforts to solve summary evaluation issues in the past two decades. NIST (National Institute of Standards and Technology) leads the effort by organizing the DUC and TAC challenges. (Huang et al., 2010) has given four pillars that should be considered to generate summaries:

- Information Coverage
- Information Significance
- Information Redundancy
- Text Coherence

In the discipline of automatic text summarization, evaluating the summary is a critical task. Evaluating the summary and increasing the development of reusable resources and infrastructure aids in comparing and replicating findings, adding competition to improve the outcomes. However, carefully evaluating many texts to acquire an unbiased perspective is impossible. As a result, accurate automatic evaluation measures are required for quick and consistent evaluation. It is difficult for people to recognize what information should be included in a summary; therefore, evaluating it is difficult. Information changes depending on the summary's purpose, and mechanically capturing this information is a challenging undertaking (Gambhir & Gupta, 2017). Evaluation of the ATS system is given in fig.7 below:

**a) Extrinsic Evaluation**: An extrinsic evaluation looks at how it influences the accomplishment of another task (Text classification, Information retrieval, Question answering). i.e., a summary is termed a good summary if it provides help to other tasks. Extrinsic evaluations have looked at how summarization affects tasks such as relevance assessment, reading comprehension, etc.

- Relevance evaluation: Various methods are used to analyse a topic's relevance in the summary or the original material.

- Reading comprehension: After reading the summary, it assesses whether it is possible to answer multiple-choice assessments.

**b) Intrinsic Evaluation**: An Intrinsic evaluation looks at the summarization system on its own. The coherence and informativeness of summaries have been the focus of intrinsic evaluations. Evaluations based on comparisons with the model summary/summaries and evaluations based on comparisons with the source document are the two types of intrinsic techniques (Steinberger & Ježek, 2009)**.**

It assesses the quality of a summary by comparing the coverage of a machine-generated and a human-generated summary. The two most significant aspects of judging a summary are its quality and informativeness. A summary's informativeness is usually assessed by comparing it to a human-made summary, such as a reference summary. There is also faithfulness to the source paradigm, which examines if the summary contains the same or similar material as the original document. This method has a flaw: how it can be determined which concepts in the document are relevant and which are not?

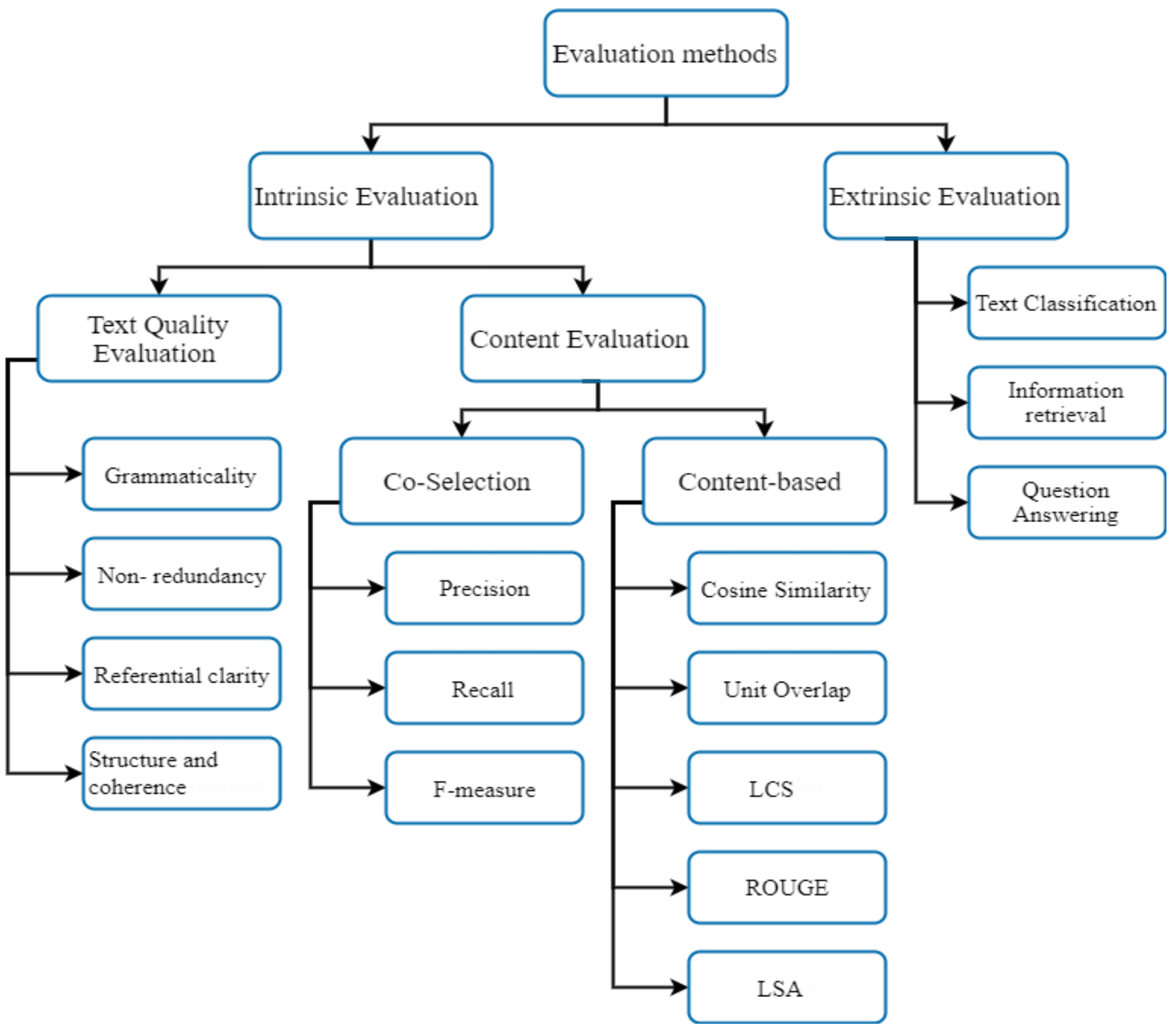

Fig. 7: The evaluation Techniques for Automatic Text Summarization

### 4.1 Content Evaluation

- ***Co-selection*:** Only identical sentences can be used in co-selection measures. It ignores the reality that even though two sentences are written differently, they can contain the same information. In addition, summaries provided by two separate authors rarely contain similar sentences. Co-selection can be calculated by precision, recall and F-measure.

a. *Precision:* Precision is the fraction of relevant instances among the retrieved instances.

$$Precision = \frac{|\{relevant\ documents\} \cap \{retrived\ documents\}|}{|\{retrived\ documents\}|} \quad (1)$$

$$Precision = \frac{True\ Positive}{True\ Positive + False\ Positive} \quad (2)$$

b. *Recall:* Recall is the fraction of relevant instances that were retrieved.

$$Recall = \frac{|\{relevant\ documents\} \cap \{retrived\ documents\}|}{|\{relevant\ documents\}|} \quad (3)$$

$$Precision = \frac{True\ Positive}{True\ Positive + False\ Negative} \quad (4)$$

c. *F-measure:* It is computed by combining recall and precision.

$$F - measure = 2 * \frac{Precision * Recall}{Precision + Recall} \quad (5)$$

- ***Content-based:*** Drawbacks of co-selection methods are handled by content-based methods.
  a. *Cosine Similarity:* Cosine Similarity can be measured as,

$$\cos(X, Y) = \frac{\sum_i x_i y_i}{\sqrt{\sum_i (x_i)^2}\sqrt{\sum_i (y_i)^2}} \quad (6)$$

Where, X and Y are representations of a system summary and its reference document based on the vector space model.

b. *Unit Overlap:* Unit Overlap can be calculated as,

$$\text{overlap}(X, Y) = \frac{\|X \cap Y\|}{\|X\| + \|Y\| - \|X \cap Y\|} \quad (7)$$

Where, X and Y are representations based on sets of words or lemmas. $\|X\|$ is the size of set X.

c. *Longest Common Subsequence (LCS):* the LCS formula is defined as shown in equation (8),

$$\text{LCS}(X, Y) = \frac{length(X) + length(Y) - edit_{di}(X, Y)}{2} \quad (8)$$

Where, X and Y are representations based on sequences of words or lemmas, LCS (X, Y) is the length of the longest common subsequence between X and Y, length(X) is the length of the string X, and $edit_{di}$(X, Y) is the edit distance of X and Y.

d. *ROUGE (Recall-Oriented Understudy for Gisting Evaluation):* It was firstly introduced by C. Lin & Rey (2001). It contains measures for automatically determining the quality of a summary by comparing it to other (ideal) summaries generated by people. The measures count the number of overlapping units such as n-grams, word sequences, and word pairs between the computer-generated summaries to be evaluated and the ideal summaries written by humans. ROUGE includes five measures like ROUGE-N, ROUGE-L, ROUGE-W, ROUGE-S and ROUGE-SU.

- ROUGE-N counts the number of N-gram units shared by a given summary and a group of reference summaries, where N is the length of the N-gram. i.e., ROUGE-1 for unigrams and ROUGE-2 for bi-grams.
- ROUGE-L calculates the LCS (Longest Common Subsequence) statistic. LCS is the maximum size of a common subsequence for two given sequences, X and Y. ROUGE-L estimates the ratio of the LCS of two summaries to the LCS of the reference summary.
- ROUGE-W is the weighted longest common subsequence metric. It is a step forward from the basic LCS strategy. ROUGE-W prefers LCS with successive common units. Dynamic programming can be used to compute it efficiently.
- ROUGE-S (Skip-Bigram co-occurrence statistics) calculates the percentage of skip bigrams shared between a single summary and a group of reference summaries. The skip bigrams are word pairs in the sentence order with random gaps.
- ROUGE-SU is a weighted average of ROUGE-S and ROUGE-1 that expands ROUGE-S to include a unigram counting unit. This is a step forward from ROUGE-S.

e. *LSA-based method:* This method was developed by Steinberger & Ježek (2009). If there are m terms and n sentences in the document, we will obtain an m*n matrix A. The next step is to apply Singular Value Decomposition (SVD) to matrix A. The SVD of an m*n matrix A is defined as given in equation (9):

$$A = U\Sigma VT \qquad (9)$$

In terms of NLP, SVD (Singular Value Decomposition) is used to generate the document's latent semantic structure, represented by matrix A: that is, a breakdown of the original document into r linearly-independent basis vectors that express the document's primary 'Topics'. SVD can record interrelationships between terms, allowing concepts and sentences to be clustered on a 'semantic' rather than a 'word' basis.

### 4.2 Text Coherence or Quality Evaluation:

a. Grammaticality: The text should not contain non-textual items (i.e., markers), punctuation errors or incorrect words.
b. Non-redundancy: The text should not contain redundant information.
c. Reference clarity: The nouns and pronouns should be referred to in summary. For example, the pronoun he has to mean somebody in the context of the summary.
d. Coherence and structure: The summary should have good structure, and the sentences should be coherent.

The linguistic characteristics of the summary are properly considered here. Non-redundancy, focus, grammaticality, referential clarity, and structure and coherence are five questions based on linguistic quality used in DUC (Document Understanding Conference) and TAC (Text Analysis Conference) conferences to evaluate summaries not needed to be reviewed against the reference summary. Expert human assessors manually score the summary based on its quality, awarding a score to the summary according to a five-point scale (Gambhir & Gupta, 2017).

The text quality of a summary can also be checked by examining several readability variables. Text quality is analysed using various criteria such as vocabulary, syntax, and discourse to estimate a correlation between these characteristics and previously acquired human readability ratings. Unigrams represent vocabulary, while the average number of verbs or nouns represent syntax.

### 4.3 Automatic Text Summarization Evaluation Programs

SUMMAC (TIPSTER Text Summarization Evaluation) was the first conference where automatic summarization systems were reviewed, and it was held at the end of the 1990s, where text summaries were assessed using two extrinsic and intrinsic criteria. DUC (Document Understanding Conferences), which took place every year from 2001 to 2007, is another notable conference for text summarizing. Initially, activities at DUC conferences like DUC 2001 and DUC 2002 featured generic summarizing of single and multiple documents, which was later expanded to include a query-based summary of multiple documents in DUC 2003. Topic-based single and multi-document cross-lingual summaries were assessed in DUC 2004. Multi-document, query-based summaries were examined in DUC 2005, and DUC 2006, whilst multi-document, update, query-based summaries were evaluated in DUC 2007. However, in 2007, DUC conferences were no longer held because they were absorbed into the Text Analysis Conference (TAC), which featured summarization sessions. TAC is a series of evaluation workshops designed to promote research in the domains of Natural Language Processing and related fields. The TAC QA program arose from the TREC QA program. The Summarization track aids in the development of methods for producing concise, cohesive text summaries. Every year TAC workshops have been held since 2008.

## 5. Frequently used Dataset for ATS

There are applications of ATS systems that are widely spread worldwide and know the available data globally. So, to perform the text summarization task essential thing is the data. Not all data can be directed feed to the system. It required prepossessing and other treatments. The machine learning-based approaches need a huge training dataset with ideal summaries to train the model. Also, the ideal or sample Dataset is needed to evaluate a particular ATS system. That sample data is manually generated or created by human researchers. The list of the Dataset available for the ATS task is very long. A very few datasets are given below:

- ***DUC:*** The National Institute of Standards and Technology (NIST) provides these datasets, the most prevalent and widely used datasets in text summarization research. The DUC corpora were distributed as part of the DUC conference's summarizing shared work. The most recent DUC challenge took place in 2007. Datasets for DUC 2001 through DUC 2007 are available on the DUC website.
- ***Text Analysis Conference (TAC) Datasets:*** DUC was added to the TAC as a summary track in 2008. To gain access to the TAC datasets, you must first fill out the application forms available on the TAC website.
- ***Gigaword:*** Created by (Rush et al., 2015), Headline-generation on a corpus of article pairs from Gigaword consisting of around 4 million articles in the English language.
- ***LCTCS:*** Created by Chen (2015), the LCSTS Dataset was constructed from the Chinese microblogging website Sina Weibo. It consists of over 2 million real Chinese short texts with short summaries given by the author of each text. Requires application in the Chinese language.
- ***wikiHow:*** Created by Koupaee & Wang (2018), the WikiHow Dataset contains article and summary pairs extracted and constructed from an online knowledge base written by different human authors in the English language. There are two features: - text: wikiHow answers texts. - headline: bold lines as summary.
- ***CNN:*** CNN/DailyMail non-anonymized summarization dataset. The CNN / Daily Mail Dataset is an English-language dataset containing just over 300k unique news articles written by journalists at CNN and the Daily Mail. The current version supports both extractive and abstractive summarization, though the original version was created for machine-reading and comprehension and abstractive question answering.

## 6. Application, Challenges and future scope

### 6.1 Applications of ATS

There are numerous uses for automatic text summarizing. As we see how text summarization is divided into many more categories. All these categories further lead us to treasure of ATS's applications. This subsection includes some of the applications of ATS system. Table 4 shows the research survey on application of ATS system. The table includes article name, the method & the dataset used in a particular article, the performance of the system proposed in that particular research study and advantages & disadvantages of that article.

Here are a few examples:

- Improving the performance of classic IR and IE systems (using a summarization system in conjunction with a Question-Answering (QA) system); (De Tre et al., 2014) (S. Liu et al., 2012) (Perea-Ortega et al., 2013)
- News Summarization and Newswire generation (Tomek,1998) (Bouras & Tsogkas, 2010)
- Rich Site Summary (RSS) feed summarization (Zhan et al., 2009)
- Blog Summarization (Y. H. Hu et al., 2017)
- Tweet Summarization (Chakraborty et al., 2019)
- Web page Summarization (Shen et al., 2007)
- Email and email thread Summarization (Muresan et al., 2001)
- Report Summarization for business men, politicians, researchers, etc. (Lloret et al., 2013)
- Meeting Summarization.
- Biographical extracts
- Legal Document Summarization (Farzindar & Lapalme, 2004)
- Books Summarization (Mihalcea & Ceylan, 2007)
- Use of Summarization in medical field (Feblowitz et al., 2011)(Ramesh et al., 2015)

Table 4: Research survey on Application of ATS System

| Citation | Article | Model/methods applied | Dataset Used | Performance | Advantages/Pros | Disadvantages/Cons |
|---|---|---|---|---|---|---|
| (Tomek,1998) | A Robust Practical Text Summarization | Multi-Document, Extractive Text summarization | New York times online news | Length of summaries: The summaries are only 5% to 10% of original text so it can be quickly read and understood. | - The algorithm is very robust that efficiently process a large range of documents, domain independent and can be easily accepted by most of the European languages.<br>- It can work in two modes: Generic and Topical<br>-Worked on DMS (Discourse Macro Structure) to conquer the shortcomings of sentence-based summarization by working on paragraph level | Quality of summaries not that much good. It can be improved by additional paragraph scoring function. |
| (Muresan et al., 2001) | Combining linguistic and machine learning techniques for email summarization | Extractive text summarization with machine learning | Emails | Precision: 83%<br>Recall:85.7% | Linguistic Knowledge Enhances Machine Learning | Deep linguistic knowledge is required. |
| (Mckeown et al., 2002) | Tracking and summarizing news on a daily basis with Columbia's Newsblaster | Multi-document summarization with extractive approach | News sites | - | These research achievements are incorporated into “Newsblaster” | Personalization of Newsblaster and restricting it to user preferred topics or questions is still having to be done |
| (Farzindar & Lapalme, 2004) | Legal Text Summarization by Exploration of the Thematic Structure and Argumentative Roles | Extractive Text summarization | Corpus contains 3500 judgements of Federal court of Canada | Preliminary results are very promising.<br>-F-measures: 0.935 on average for all stages | -Summary presented as table style<br>-focused on many categories of judgements i.e., Copyright, Air low, human rights etc. | The System is not properly for many other remaining categories of judgements |
| (Mihalcea & Ceylan, 2007) | Explorations in automatic book summarization | Extractive text summarization with TEXTRANK approach | gold standard” data set of 50 books | F-Measure: 0.404 | it introduced a new summarization benchmark, specifically targeting the evaluation of systems for book summarization | Exhaustive method when we have short length book |
| (Ling et al., 2007) | Generating gene summaries from biomedical literature: A study of semi-structured summarization | Multi-Document, Extractive Text summarization | test set with 20 genes | ROUGE-N performs better than existing methods | Semi- structured summaries are generated which consist of sentences regarding specific semantic aspects of a gene. | - High quality of data required. (Data dependent)<br>-Redundant information in generated summary |
| (Shen et al., 2007) | Noise reduction through summarization for Web-page classification | Extractive Text summarization | 2 million Web pages crawled from the LookSmart Web directory | F-measures for hybrid approach (supervised and unsupervised):<br>Naïve Byes- 72.0 ± 0.3<br>SVM- 72.9 ± 0.3 | Removing noise from web pages while preserving most relevant features to increase accuracy of web classification | - Focus only isolated web pages. Does not include hyperlinks. |
| (Zhan et al., 2009) | Gather customer concerns from online product reviews – A text summarization approach | Extractive Text summarization | Five datasets from Hu’s corpus (M. Hu & Liu, 2004) and 3 sets from Amazon | Average responsiveness scores: 4.3 | find and extracts important topics from a set of online reviews and then ranks these retrieved topics | - For different online sites the style of reviews is written in differently. This difference is not efficiently entertained. |
| (Bouras & Tsogkas, 2010) | Noun retrieval effect on text summarization and delivery of personalized news articles to the user's desktop | Multi-Document Extractive Text summarization | Numerous news portals around the internet | Precision is boosted by using noun retrieval effect.<br>With new personalization scheme increase to around 17% for precision and 14% for recall. | -Enhance personalization algorithm with help of various features extracted from user’s profile and viewed history of articles.<br>- A stable system for day-to-day use is build named as PeRSSonal. | Language dependent |
| (Feblowitz et al., 2011) | Summarization of clinical information: A conceptual model | Extractive Text summarization | Day to day clinical data | - | Both computer supported and computer independent clinical tasks are analysed. | Not standardised and optimized clinical summary |
| (Lloret et al., 2011) | Text Summarization Contribution to Semantic Question Answering: New Approaches for Finding Answers on the Web Elena | Extractive Text summarization | Most relevant 20 documents from google search engine for particular Question. | Query-focused summaries gives 58% improvement in accuracy than generic summaries. shorter summaries gives 6.3% improvement in accuracy than long summaries. | Efficiently combines text summarization for semantic question answering while focusing on query-based summaries rather than generic summaries. | The summary for a particular question’s answer was built from only 20 documents which were retrieved from google search engine. |
| (S. Liu et al., 2012) | TIARA: Interactive, topic-based visual text summarization and analysis | Multi-Document Extractive Text summarization | IBM employees’ Email | Improvement in TIARA with regarding to usefulness and satisfaction of summary than TheMail. (Viégas et al., 2006) | -An Interactive visual text analysis tool TIARA which produces a visual summary of text analytic results automatically and enhanced with significant, time-sensitive topic-based summary method. | Application – specific features are not added in the tool. |
| (Kavila et al., 2013) | An automatic legal document summarization and search using hybrid system | Extractive text summarization | corpus presently consists of 100 legal documents related to criminal and civil collected | 90% results in automatic search | Hybrid system of Keyword/ Key phrase matching technique and Case based technique | Future work will a hybrid system to find the expected judgement for a new case from the past cases. |
| (Lloret et al., 2013) | COMPENDIUM : A text summarization system for generating abstracts of research papers | Hybrid Text summarization | 50 research papers from journal of medicine | -quantitative and qualitative evaluation approach<br>-Precision: 40.53 Recall:44.02 | -COMPENDIUM produces summaries for biomedical research papers automatically with both approaches: extractive and abstractive.<br>-abstractive are far better as per user perspective. | Extractive summaries generated by system are contain less similar information to human written summaries |
| (Perea-Ortega et al., 2013) | Application of text summarization techniques to the geographical information retrieval task | Extractive Text summarization | 169,477 documents collections of stories and newswires | GeoCLEF: Compression rate ranging from 20% to 90%<br>Recall<br>MAP | Two types of summaries are generated (generic & geographical) with improvement in recall and compression rates. | - approach can be only applied to single document<br>-discarding all the sentences which do not contain any geographical information may leads to loss of information. |
| (Sankarasu bramaniam et al., 2014) | Text summarization using Wikipedia | Single and Multi-Document Extractive Text summarization | DUC (2002)-567 English news articles | ROUGE-1: 0.46<br>ROUGE-2: 0.23 | -Works for both single document and multi document<br>- For selecting summary phrases, created a bipartite sentence–concept graph and presented an iterative ranking algorithm. | Evaluation is done against human-generated summary which naturally favours the leading line sin new article. |
| (Ramesh et al., 2015) | Figure-Associated Text Summarization and Evaluation | Extractive Text summarization | 94 annotated figures selected from 19 different journals | F1 score of 0.66 and ROUGE-1 score of 0.97 | Unsupervised FigSum+ system that automatically fetches associated texts, remove repetitions, and generate text summary for a particular figure. | -It requires annotated figures.<br>- Experiments results are generated by only 94 figures. |
| (Y. H. Hu et al., 2017) | Opinion mining from online hotel reviews – A text summarization approach | Machine learning based Extractive Text summarization | Reviews of Red Roof Inn and Gansevoort Meatpacking Hotel selected from TripAdvisor. com. | Mean, standard deviation and significance level. | -Proposed technique fetches top-k informative sentences from online reviews<br>-Proposed method focused on critical factors such as usefulness of a review, credibility of authors, review time, conflicts in reviews. | - TripAdvisor.com contains reviews in many more languages but the proposed approach works for only English.<br>- Small sample of participants with same backgrounds.<br>- Experiments are done for only 2 hotels. |
| (Lovinger et al., 2017) | Gist: general integrated summarization of text and reviews | Extractive text summarization with optimization-Based approach | Movie reviews, news articles | Average F-measure: 0.276<br>Average running time: 0.067235 | High F-score compared to TextRank and LexRank | High computational time and cost |
| (Chakrabor ty et al., 2019) | Tweet summarization of news articles: An objective ordering-based perspective | Extractive text summarization with LEXRANK approach | News article dataset and tweet dataset | Rouge-1, F1-score: 0.664<br>Rouge-2, F1-score: 0.548 | Capturing the diverse opinions helps in better identification of the relevant tweet set. | classification of opinion from tweets are not implemented |
| (Kumar & Reddy, 2019) | Factual instance tweet summarization and opinion analysis of sport competition | Extractive text summarization | IPL 2017 challenger 1, eliminator, challenger 2, and final competition with each having 10,000 tweets | Sentiment classifier time: -Logistic regression<br>Train: 0.27s<br>Test: 0.025 | classification of sub-events of tweets | -opinion classification is not derived. |

### 6.2 Challenges and future scope

While generating an automated text summary, one faces a lot of challenges. The first challenge is defining what constitutes a decent summary, or more precisely, how a summary might be constructed. Our requirements for a summary provide good clues as to what it should be: extractive or abstractive, general, or query-driven, etc. Even if we figure out how humans normally summarise, putting it into practice will be difficult. Creating a powerful automatic text summarizer necessitates many resources, either in terms of tools or corpora. Another challenge is summary in formativeness; how can a machine emulate human people when it comes to summarising? One of the long-standing issues has been the coherence of the summary. The shortage of resources is another most challenging problems in ATS.

Compared to the past, there are numerous powerful tools for stemming, parsing, and other tasks available now. Despite this, determining which ones are appropriate for a particular summarization problem is the issue. Furthermore, annotated corpus for ATS can be considered a challenge. The evaluation process is also a significant difficulty. Both intrinsic and extrinsic evaluation approaches were explored in this work. The language shared by a machine-generated reference summary is generally the focus of current intrinsic evaluation methods. Intuitive evaluation can create new ways to evaluate the summary based on the information it includes and how it is presented. The process of evaluation is highly subjective. First, a reasonable criterion must be defined to understand what is important and what is not. It is also unclear whether this procedure can be fully automated.

Text summarising has been around for more than fifty years, and the academic community is very interested in it; therefore, they continue to improve existing text summarization ways or invent new summary approaches to provide higher-quality summaries. However, text summarization performance is still average, and the summaries created are not perfect. As a result, by merging this system with other systems, it can be made more intelligent, allowing the combined system to perform better.

- **Conclusion**

Automatic Text summarization reduces the size of a source text while maintaining its information value and overall meaning. Automatic Text summarization has become a powerful technique for analysing text information due to a large amount of information we are given and the growth of Internet technologies. The automatic summarization of text is a thriving-known task in natural language processing (NLP). Automatic text summarization is an exciting research area, and it has a treasure of applications. This paper aims to make readers understand automatic text summarization from ground level and familiarise them with all detailed types of ATS systems. After that all different types are distinguished deeply and clearly in this study. The summarization task is mainly divided into extractive and abstractive. The study shows numerous techniques for extractive summarization, but the summaries generated by extractive summarizers are far from human-made summaries. On the other hand, abstractive summarizer is close to human summaries but not practically implemented with high performance. The combination of both extractive and abstractive is hybrid text summarization. This paper includes research survey on Extractive, Abstractive and Hybrid Text Summarization. Also, this survey article tried to cover all major application areas of ATS system and provided detailed survey on the same. There are so many methods to evaluate summarizing system and generated summaries that are included in this paper. Further it gives brief idea about frequently used datasets, conferences and programs that held every year for automatic text summarization system.

. The future study is to build a robust, domain and language independent extractive text summarization that works well with multi-documents. Similarly, because the quality evaluation of the summary is done manually by experienced assessors, it is highly subjective. There are specific quality assessment criteria, such as grammaticality and coherence, but the results are different when two experts evaluate the same summary.

**Conflicts of Interest:** On behalf of all authors, I state that there is no conflict of interest.